\documentclass[acmtog, screen]{acmart} 

\definecolor{purple}{cmyk}{0.45,0.86,0,0}
\definecolor{bleudefrance}{rgb}{0.19, 0.55, 0.91}
\definecolor{darkorange}{rgb}{1, 0.55, 0}
\definecolor{limegreen}{rgb}{0.2, 0.8, 0.2}

\def \eg {{\emph{e.g}.\thinspace}, }
\def \ie {{\emph{i.e}.\thinspace}, }

\DeclareMathOperator*{\argmax}{\arg\!\max}

\newlength\paramargin
\newlength\figmargin
\newlength\secmargin
\newlength\figcapmargin

\long\def\ignorethis#1{}

\usepackage{hyperref}

\usepackage{graphicx}
\graphicspath{{figure},{images},{example}}
\DeclareGraphicsExtensions{.png,.PNG,.jpeg,.JPEG,.jpg,.JPG,.bmp,.BMP}
\usepackage{subcaption}
\usepackage[export]{adjustbox}
\usepackage{float}
\usepackage[]{caption}	
\usepackage{lscape} 
\usepackage{overpic}
\usepackage[ruled,linesnumbered]{algorithm2e}

\SetAlFnt{\small}
\SetAlCapFnt{\small}
\SetAlCapNameFnt{\small}
\SetAlCapHSkip{0pt}
\IncMargin{-\parindent}

\usepackage{booktabs}  
\usepackage{multirow}
\usepackage{paralist}

\usepackage{mathtools}
\usepackage{amsmath}   
\usepackage{units}
\usepackage{color}

\usepackage{comment}
\usepackage{bbm}
\usepackage{soul}
\usepackage{wrapfig}

\usepackage{array}
\usepackage{tabularx}
\usepackage{tikz}
\usepackage{pgfplots} 
\pgfplotsset{compat=1.18}
\usepackage{scalefnt}
\graphicspath{{./images/}}
\usepackage{color,hyphenat,balance,booktabs}
\usepackage{graphicx}
\usepackage{xcolor}
\usepackage{overpic}
\usepackage{autonum}
\usepackage{wrapfig}
\usepackage{physics}
\usepackage{amsmath}

\SetCommentSty{mycommfont}
\usepackage{enumitem}
\newtheorem*{definition*}{Definition}
\theoremstyle{remark}

\setcitestyle{square}

\acmSubmissionID{}

\newcommand{\name}{\textit{StructRe}\xspace}

\begin{document}

\title{\name: Rewriting for Structured Shape Modeling}

\author{Jiepeng Wang}
\affiliation{
    \institution{The University of Hong Kong}
    \country{China}
}
\authornote{Work done during internship at Microsoft.}
\email{jiepeng@connect.hku.hk}
\affiliation{
    \institution{Microsoft Research Asia}
    \country{China}
}
\author{Hao Pan}
\affiliation{
    \institution{Tsinghua University}
    \country{China}
}
\email{haopan@tsinghua.edu.cn}
\affiliation{
    \institution{Microsoft Research Asia}
    \country{China}
}
\authornote{Corresponding author.}

\author{Yang Liu}
\affiliation{
    \institution{Microsoft Research Asia}
    \country{China}
}
\email{yangliu@microsoft.com}
\author{Xin Tong}
\affiliation{
    \institution{Microsoft Research Asia}
    \country{China}
}
\email{xtong.gfx@gmail.com}
\author{Taku Komura}
\affiliation{
    \institution{The University of Hong Kong}
    \country{China}
}
\email{taku@cs.hku.hk}
\author{Wenping Wang}
\affiliation{
    \institution{Texas A\&M University}
    \country{United States of America}
}
\email{wenping@tamu.edu}
%

\begin{teaserfigure}
  \centering
  \includegraphics[width=0.92\linewidth]{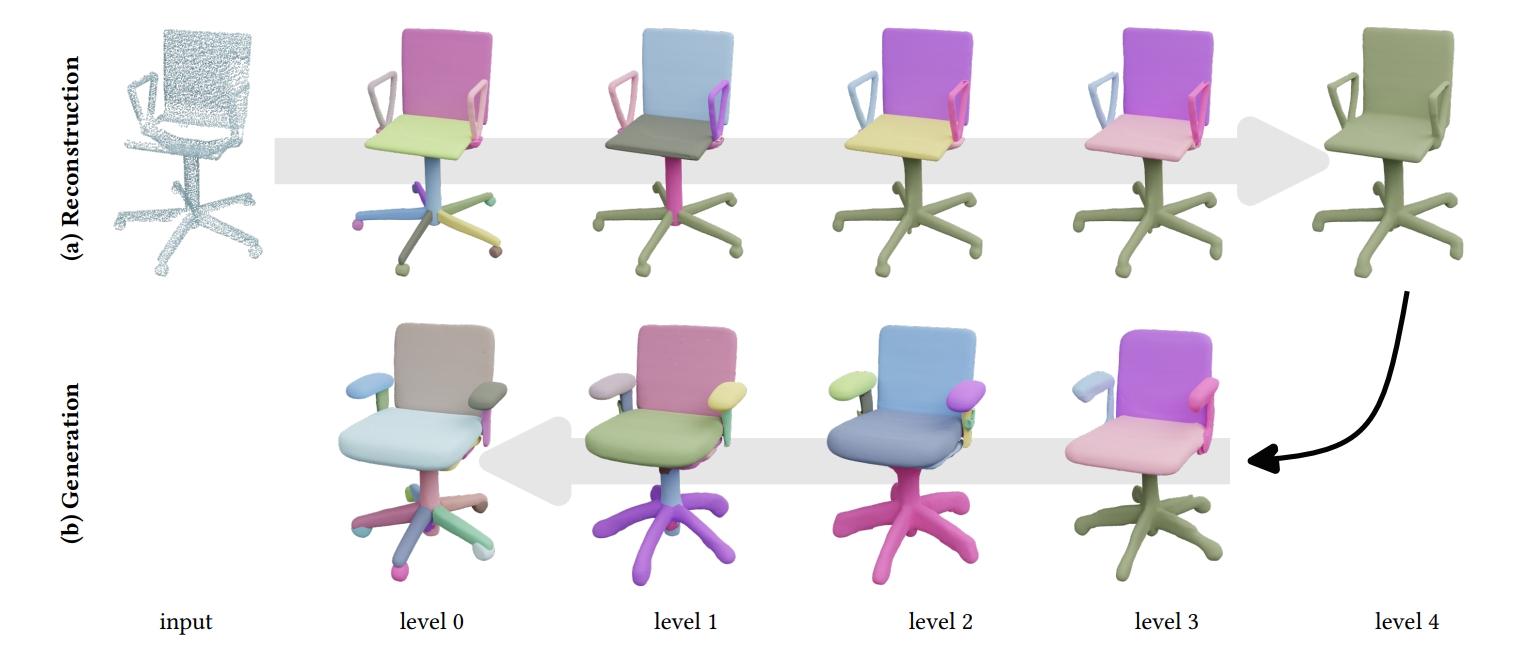}
  \vspace{-14pt}
  \caption{\textbf{Rewriting for structured shape modeling.} Parts are colored randomly. 
  \name is a structure rewriting system that can be used for (a) Structured reconstruction: given a partial scan, it inductively reconstructs the hierarchically structured shapes, by  detecting primitives from point cloud (level 0), and iteratively rewriting simpler parts into more compact parts (levels $0{\rightarrow} 4$).
  (b) Structured generation: starting from a reference shape, it deductively performs shape perturbation and downward rewriting (levels $4{\rightarrow} 0$, Sec.~\ref{subsec:comparison_gen}) to generate novel hierarchically structured shapes.
  }
  \label{fig:teaser}
\end{teaserfigure}

\begin{abstract}

Man-made 3D shapes are naturally organized in parts and hierarchies; such structures provide important constraints for shape reconstruction and generation.
Modeling shape structures is difficult, because there can be multiple hierarchies for a given shape, causing ambiguity, and across different categories the shape structures are correlated with semantics, limiting generalization.
We present \name, a structure rewriting system, as a novel approach to structured shape modeling. 
Given a 3D object represented by points and components, \name can rewrite it upward into more concise structures, or downward into more detailed structures; by iterating the rewriting process, hierarchies are obtained.
Such a localized rewriting process enables probabilistic modeling of ambiguous structures and robust generalization across object categories.
We train \name on PartNet data and show its generalization to cross-category and multiple object hierarchies, and test its extension to ShapeNet. 
We also demonstrate the benefits of probabilistic and generalizable structure modeling for shape reconstruction, generation and editing tasks.

\end{abstract}



\maketitle

\section{Introduction}

\begin{figure}
    \centering
    \includegraphics[width=0.95\linewidth]{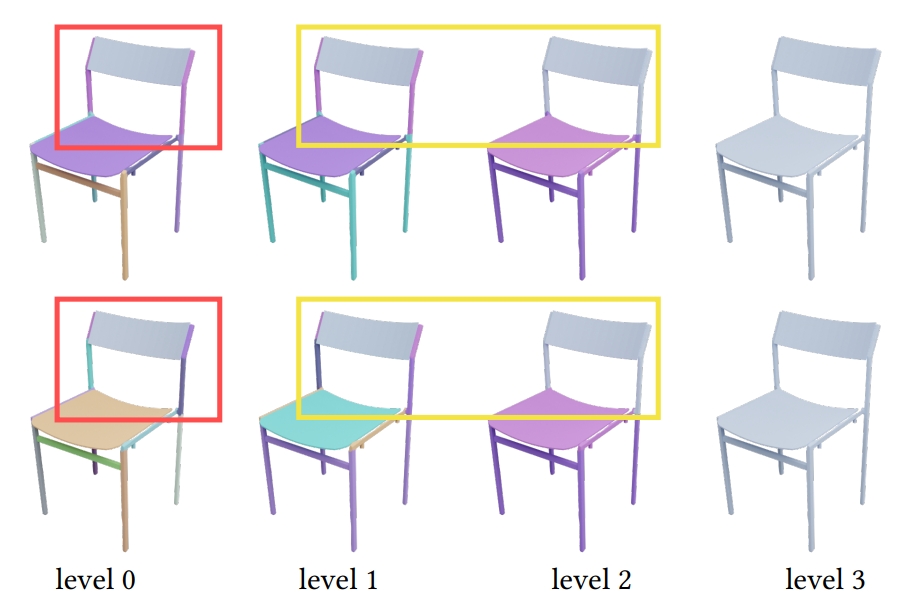}
      \vspace{-12pt}
    \caption{
    \textbf{Shape structure ambiguity}. A single shape can have multiple possible decompositions. As highlighted by the red rectangles, the chair back can be decomposed into 2 parts or 5 parts at the same level (level 0), with similar variations observed at level 1 and 2 highlighted by the yellow rectangles, illustrating the structural ambiguity inherent in shape decomposition. 
    }
    \label{fig:ambiguity_chair_table}
\end{figure}

Man-made objects generally have structures, i.e., hierarchies of parts, which decompose geometry modularly and provide the foundation for shape recognition, analysis, and modeling \cite{PartsOfRecog,Mitra2013Survey,Mo2019partnet}.
While heuristic approaches mainly rely on geometric concavities to build such structured representations \cite{SegSurvey08,SegSurvey18}, part and hierarchy recognition is deeply related to semantics, and therefore data-driven deep learning approaches have made great advances in modeling the proper structures \cite{li2017grass,mo2019structurenet,wu2019sagnet,wu2020pqnet,Mo2020StructEdit,Wang2022SlotMachine,Petrov2022ANISE,Yang2022DSGNet}.
These works generally model the structures as hierarchical trees, and aim to generate the comprehensive structure deterministically within one pass. 
However, there can be multiple plausible structures for a single object \cite{Mo2019partnet,greff2020binding} 
(\eg see Fig. \ref{fig:ambiguity_chair_table}, where the chair back can be decomposed into 2 or 5 parts according to the interpretations of different users); such ambiguity seriously challenges the deterministic structure generation. 
To address this issue, existing learning-based methods rely on object category priors to select certain structures among others.
Nevertheless, since categorical priors encode semantics of parts and hierarchies, models bound by these priors do not generalize beyond their specific training categories.

\begin{figure}
    \centering
      \includegraphics[width=0.95\linewidth]{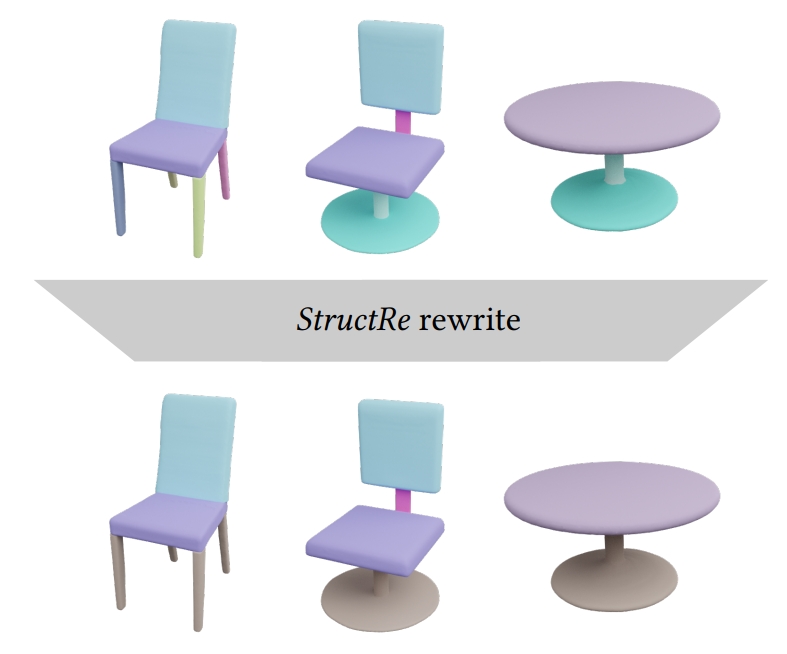}
      \vspace{-12pt}
    \caption{\textbf{Similarity of local composition across shapes.} From top to bottom: three objects have their bases detected and grouped into single parts (in brown) by a common query token of our rewriting network (c.f. Sec.~\ref{subsec:rewrite_network}). 
    The rewriting system learns such local compositions for resolving structure ambiguity and achieving robust generalization across object categories. 
    }
    \label{fig:observation} 
\end{figure}

In this paper, instead of producing a complete structure in one pass, we take a local and probabilistic approach for structured shape modeling (Fig.~\ref{fig:teaser}). 
First, we note that although complete objects are made of vastly different combinations of parts at different levels, within localized parts and neighboring levels, the composition patterns recur across objects (Fig.~\ref{fig:observation}).
Based on this observation, we propose to learn a rewriting system to break down shape structure modeling into more tractable steps that generalize better.
Specifically, in each step, the rewriting system is trained to organize local parts into a larger component, or to decompose compact components into detailed parts, and allow for sampling from the different combinations that are equally plausible \cite{greff2020binding} (Fig.~\ref{fig:sampling}).
Therefore, the ambiguity of different structures for a given geometry is resolved by concurrent probabilistic sampling of different rewriting rules. 
By focusing on local structural patterns rather than being bound globally by categorical priors, the rewriting system generalizes across shape categories (Sec.~\ref{subsec:result_analysis}).

Learning a rewriting system for local structure composition faces several challenges. 
First, object parts of different categories and granularities must be encoded compactly and uniformly for rewriting learning. 
Second, the rewriting task faces the dilemma of context sensitivity and locality: given a set of object parts, the rewriting system must group the relevant parts to compose/decompose, while not interfered by parts across groups, for robust generalization across object categories.
Finally, in order to handle ambiguous structures, the system should generate diverse structures probabilistically.

We present \name, a framework that addresses the above challenges and implements a local and probabilistic rewriting system for structured shape modeling.
We represent an object part dually as a latent code anchored within a spatially positioned bounding box, and as a high-resolution implicit function defined through a decoded feature grid (Fig.~\ref{fig:pipeline}(a)(b)).
To enable the reasoning over object parts of diverse granularities, we build a unified latent space of part shapes with vastly different complexity, ranging from simple parts like bars and cylinders to complex objects in their entirety (Fig.~\ref{fig:pipeline}(b)).
Given point clouds and part codes, we learn the structure rewriting network that transitions among parts to detect and compose/decompose them, according to specified upward/downward directions (Fig.~\ref{fig:pipeline}(c)). 
Due to the simple syntax of rewriting models that replace input parts with output ones, the rewriting network can be implemented by a pair of transformer-based part set encoder and decoder.
To achieve both context sensitivity and locality, the transformer encoder parses input part codes with global attention and detects the context-sensitive patterns;
furthermore, we generate random combinations of composition patterns to train the network, so that it is robust to interference of unrelated parts and generalizable to novel compositions.
To model ambiguous solutions, we use an iterative decoding scheme for the decoder (Fig.~\ref{fig:pipeline}(c)) that samples diverse but consistent structures according to probability.

We train and test \name on the large-scale PartNet dataset \cite{Mo2019partnet} containing structure annotations.
Extensive experiments show that our model can reconstruct and generate structured shapes across object categories and multiple objects (Figs.~\ref{fig:teaser},\ref{fig:small_categories},\ref{fig:multiple_objects}, etc.).
Furthermore, we validate its extension to ShapeNet \cite{chang2015shapenet}, where it infers meaningful structures in zero-shot or through few-shot tuning (Figs.~\ref{fig:shapenet_hierarchy},\ref{fig:shapenet_parts},\ref{fig:vis_recon_few_shot}). 
Finally, structure rewriting enables highly regularized shape modeling, including reconstruction, generation, editing, etc. (Sec.~\ref{sec:results}).

To summarize, we make the following contributions:
\begin{itemize}[leftmargin=*]\setlength\itemsep{0mm}

    \item We propose a rewriting formulation for hierarchically structured shape modeling that breaks down the ambiguous task into local composition learning.
    
    \item We implement the rewriting system by \name consisting of neural networks and training mechanisms that handle context awareness and generalization simultaneously, and support probabilistic sampling of different structures. 
    
    \item We show that \name enables robust structured shape modeling across object categories and extending to novel categories with zero or few shots.
    
\end{itemize}
Code and data will be released to facilitate future research.

\section{Related Work}

\paragraph{Structured shape modeling}

Structured shape representations encode a shape as a hierarchy of parts. By decoupling the shape structure from part geometry, structured shape representation not only matches the way humans perceive objects in cognition \cite{PartsOfRecog,Biederman1987recognition,Hochstein2002ViewFT}, but also enables computationally effective shape analysis and modeling \cite{Mitra2013Survey}. 
While early efforts discover structured representations by low-level geometric cues like tubular structures and curvature extrema  \cite{BiasottiMatching03,Mortara04}, it is noted from the very beginning that shape structures are derived from both geometric cues and semantics of decompositions. 
Therefore, structured shape representation obtains its most significant progress when data-driven machine learning techniques are applied in combination with geometric analysis.

Most of the previous works on structured shape representation learning have followed an auto-encoding scheme, and identify the encoding/decoding of both discrete structures and continuous part geometry, as well as their complex interactions, by neural networks as the major challenge.
For example, given a predefined binary tree hierarchy built by symmetry and proximity, GRASS \cite{li2017grass} proposes a recursive network to collapse lower-level node features into higher-level ones until the root node feature is obtained, and inverts the encoding process with a mirrored recursive decoder that expands the root node feature into a tree that is compared with the input tree for supervision.
In contrast, SAGNet \cite{wu2019sagnet} learns the auto-encoding of geometry and structure in two streams, which are additionally intertwined by an attention mechanism for joint distribution modeling. 
Later, PartNet \cite{Mo2019partnet} provides a large-scale dataset containing human user annotations of object structure hierarchies, on which more sophisticated networks can be trained and tested.
For example, StructureNet \cite{mo2019structurenet} presents an autoencoder of shape structures based on a graph neural network, such that input hierarchies with $n$-ary trees can be processed robustly.
Subsequently, numerous approaches have improved in terms of part geometry generation \cite{gao2019sdmnet,Petrov2022ANISE}, structure and geometry disentanglement \cite{Yang2022DSGNet}, structure generation with more flexible sequential networks \cite{wu2020pqnet}, and tree-conditioned point cloud generation \cite{mo2020pt2pc}.
Furthermore, \cite{RobertsLSD2021} enables level-of-detail generation that preserves high-level hierarchy but varies lower-level structures by probabilistic subtree decoding. 
Additionally, building on StructureNet, Seg\(\& \)Struct \cite{kim2023segstruct} introduces a framework that harnesses the synergy between segmentation and structure inference to enhance the performance of both tasks.

While previous works have learned powerful neural networks for structured shape representation, the structures have been mostly learned by auto-encoding in one pass, which faces the inherent ambiguity of multiple plausible structures for a given shape.
To mitigate this issue, these works exploit template semantics of each object category and train class-specific models that do not generalize across or to unseen categories.
We instead present a local probabilistic rewriting model for structure modeling, which handles the structure ambiguity problem by probabilistic rewriting, and enables generalization across categories by rewriting locally.

\paragraph{Rewriting systems}
As formal and general models of computation, various rewriting systems, e.g., string rewriting, L-system and graph rewriting, have been developed for applications like code analysis and optimization. 
We refer the readers to \cite{Meseguer20years,Dershowitz1990RewriteS} for comprehensive discussions of rewriting systems.
In computer graphics, procedural models like L-systems have been studied extensively for shape generation \cite{MullerProcedural06,TaltonProcedural11}.
Recently, there is a surge of inverse procedural models that use deep learning-based techniques to recover model parameters and shape programs from raw geometry input \cite{EGSTARNeuroSymbolic,Guo20Lsystem}, with which our work is closely related.
For example, ShapeCoder \cite{jones2023shapecoder} aims to discover a set of explicit abstraction functions and programs to explain input shapes. However, it primarily focuses on structures represented as cuboids, ignoring geometric details within each cuboid.
Most of these works first design a domain specific language (or grammar), and then devise pipelines including neural networks and optimization procedures to compose instructions within the language that correspond to input unstructured data.

In this paper, we learn a rewriting system for structured shape modeling, based on the observation that locally the structural similarity across shapes can be accurately captured by a rewriting model.
The end-to-end system learns both inductive and deductive rewriting (Fig.~\ref{fig:teaser}), over not only raw point cloud but also structured parts.
Our network design features both probabilistic sampling and local rewriting, and applies to structured shape modeling with robustness and generalization, in comparison with category-specific tree-structured models or the straightforward serialization of structures by auto-regressive models (Sec.~\ref{subsec:comparison_recon}).
The contrast shows that a neural-symbolic model founded on the generality of rewriting can be more powerful in terms of both robustness and flexibility.

\section{Structure Modeling by Rewriting}
\label{sec:overview}
\begin{figure*}
  \centering
  \includegraphics[width=\linewidth]{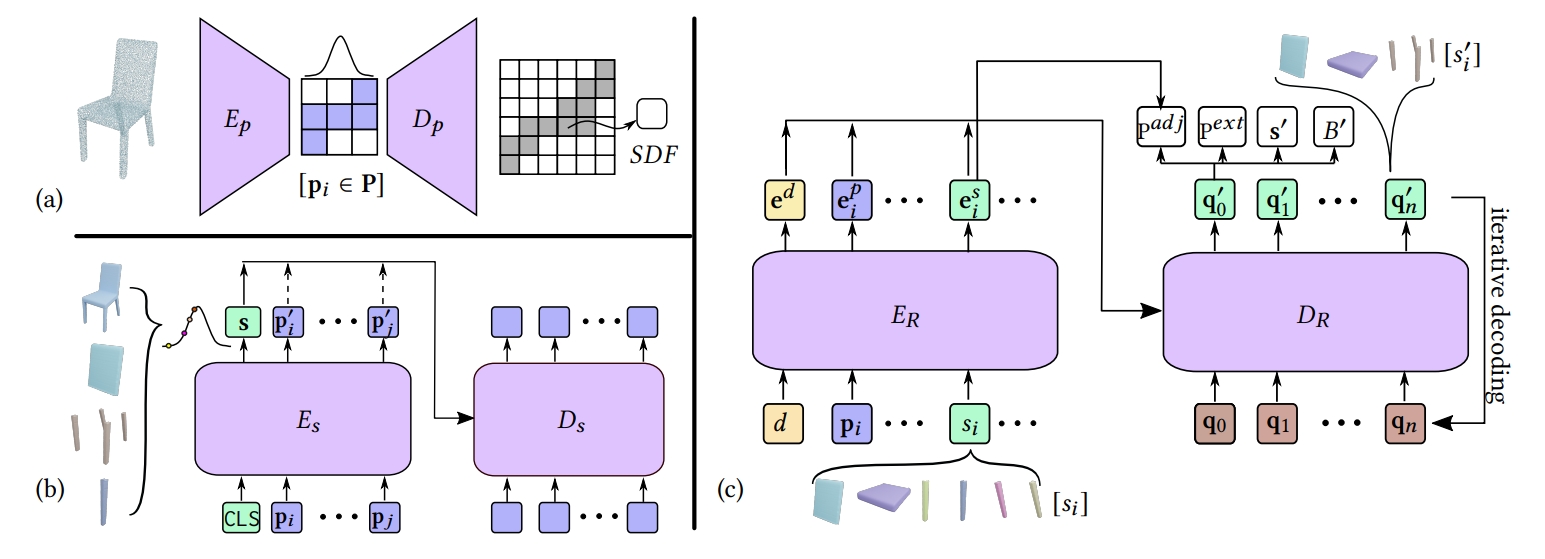}
  \vspace{-12pt}
  \caption{\textbf{Pipeline overview.} \name consists of three modules that are trained sequentially. (a) first, we learn a compact patch encoding $\vb{p}_i\in\vb{P}$ of detailed geometry by training an autoencoder $(E_p,D_p)$. (b) second, we learn a part shape encoding $\vb{s}\in\vb{\Sigma}$ by training an autoencoder $(E_s,D_s)$ on all parts and shapes, which crucially provides a unified space over which to reason rewriting. (c) third, we learn the rewriting model by training $(E_R,D_R)$ that transits from the input parts $[s_i]$ and patches $[\vb{p}_i]$ to the output parts $[s'_i]$, controlled by the direction $d\in \{\uparrow,\downarrow\}$. The decoder $D_R$ features iterative decoding that refines the output from a previous iteration subsequently. In the illustration, four chair legs are rewrote into a base.
  }
  \label{fig:pipeline}
\end{figure*}

Rewriting systems have been widely used to formalize computation semantics with simple syntax \cite{dershowitz1985computing-rewriting}. 
Given an input string and a set of \emph{rewriting rules}, these systems repeatedly recognize in the input the left-hand side (LHS) terms of applicable rules and replace them with their corresponding right-hand side (RHS) terms, achieving state transition. 
The process is very similar to the hierarchical modeling of shape structures: starting from primitive parts, one can iteratively detect the higher-level parts grouping the primitive parts and replace with them, until the complete shape is recognized as one part; the modeling can also be applied in a reverse direction, i.e., gradually decomposing higher-level parts to lower-level ones. 
This rewriting perspective to shape structure modeling is local by definition, and probabilistic by allowing concurrent rules that have the same LHS patterns but different RHS terms \cite{Meseguer20years} (c.f. Fig.~\ref{fig:sampling}).
Local and probabilistic rewriting, when applied to shape structure modeling, overcomes the structure ambiguity problem and enables robustness and generalization across object categorical semantics.
Moreover, the simple syntactical format of rewriting rules permits implementation by simplified network architectures (Fig.~\ref{fig:pipeline}), in contrast to networks \cite{mo2019structurenet,li2017grass} that are tightly bound to hierarchical data structures.

We formulate the hierarchical structure modeling of shapes as a rewriting model:
\begin{equation}
    \mathcal{R} = (\Sigma, R),
    \label{eq:rewrite}
\end{equation}
where $\Sigma, R$ are defined as:
\begin{itemize}[leftmargin=*]\setlength\itemsep{0mm}
    \item  $\Sigma$: the collection of geometry shapes, covering shape parts of all levels, from primitives (e.g. bars and cylinders) to higher-level parts related to semantics (e.g, a chair base composed of four legs) and even complete shapes (e.g. a chair). 
    The unified collection lays the foundation for reasoning over compositional structure relations, as we can directly address shapes of different complexities within a single rewriting step.
    Furthermore, we denote all combinations of shapes and parts as the set $\Sigma^*$.
    \item  $R$: a set of rewriting rules $r: \Sigma^*\times D\rightarrow \Sigma^*$ from instantiated parts $[s_i] \subset \Sigma^*$ to another set of parts  $[s'_i]\subset \Sigma^*$ according to the direction $d\in D$, such that elements $s_i$ and $s'_j$ form parent-child relationships. $D = \{\uparrow,\downarrow\}$ specifies the directions of the rewriting process: $\uparrow$ means that elements of $[s'_i]$ are higher-level parts grouping elements in $[s_i]$, and vice versa for $\downarrow$.
\end{itemize}
To learn the rewriting system $\mathcal{R}$,
we implement it by learning \name consisting of neural networks, to achieve robustness and generalization against shape variations. 
Fig.~\ref{fig:pipeline} gives an overview of the three modules of \name framework.

First, following common practice in high-fidelity latent representation training for images and shapes \cite{van2017vqvae,dosovitskiy2020vit,rombach2022stablediffusion,yan2022shapeformer}, we build a regularized latent space $\vb{P}$ of patch codes for local geometry encoding, by training a variational autoencoder $(E_p,D_p)$ such that a high-resolution shape can be compactly encoded into a sequence of patch codes $[\vb{p}_i\in \vb{P}]$ (Sec. \ref{sec:svqvae}). 

Second, we build a regularized latent space $\vb{\Sigma}$ for object part encoding. In particular, building on patch codes, we train a transformer-based autoencoder $(E_s, D_s)$ to encode all possible shapes and parts into single codes $\vb{s}\in \vb{\Sigma}$ (Sec. \ref{sec:unistruct}), which provides the foundation for rewriting rules definition. 

Finally, given the latent code representations of geometry patches and object parts, we design an encoder-decoder network $R=(E_R,D_R)$ that implements the rewriting rules by transitioning between different samples composed from $\vb{\Sigma}$ and $\vb{P}$ that form part hierarchies (Sec.~\ref{sec:structmapping}). 
In particular, we instantiate the normalized shape parts with scales and 3D positions, but omit the rotations of a part, effectively treating rotated versions as different shapes.

To provide a clear understanding of the entire rewriting process, we illustrate the process for an example shape with step-by-step input/output details in Fig. 3 of the supplemental.

\section{Shape space representation learning}
\label{sec:shape_rep_learning}

We learn a latent shape space to provide the domain for defining rewriting rules that transit between parts of different levels (Sec.~\ref{sec:structmapping}).
Learning the shape space involves mapping diverse shapes and parts to regularized and compact latent codes, supervised by decoding the latent codes to detailed and high-quality shapes.
We build on state-of-the-art techniques to learn such latent spaces through variational auto-encoders, and make adaptations to preserve shape quality.

\subsection{Learning patch-based geometry encoding}\label{sec:svqvae}

To efficiently encode 3D shapes represented by 3D point clouds into compact code sequences, we adopt a sparse VAE $(E_p,D_p)$, where both the encoder $E_p$ and decoder $D_p$ are sparse CNNs \cite{choy2019sparsecnn}. 
The decoder $D_p$ produces sparse feature grids that can be interpolated and further decoded to the implicit SDF representations of input shapes, for the sake of high-quality shape reconstruction.

Specifically, given an input 3D shape represented by points $[\vb{x}^{p}_i\in \mathbb{R}^3]$ and normal vectors $[\vb{x}^{n}_i\in \mathbb{R}^3]$, we first quantize the points into a sparse grid with $256^3$ resolution, then feed the quantized voxels with averaged position and normal vectors into the encoder $E_p$.
$E_p$ gradually downsamples the voxels into a sparse feature grid $[\vb{p}_{i}\in \vb{P}]$ of resolution $8^3$ at the bottleneck between $E_p$ and $D_p$. The latent space $\vb{P}$ of dimension 256 is shared among patch codes. 
To ensure that patch codes $\vb{p}_{i}$ encode shape variations locally, similar to previous works \cite{mittal2022autosdf,yan2022shapeformer,rombach2022stablediffusion}, we use a shallow network for the encoder $E_p$ such that each latent code has a small receptive field perceiving local patches only. 
The decoder $D_p$ reconstructs the shape in two steps: 1) gradually upsampling $[\vb{p}_{i}]$ to a sparse feature grid of resolution $128^3$, and 2) decoding to SDF values of sample points around the shape surface by feature interpolation and a shared MLP decoder. 
To extract the shape surface, we use marching cubes \cite{MarchingCubes87} to search for the zero-level set of SDF on non-empty voxels at resolution $256^3$. 

The regularity of the latent space $\vb{P}$ is crucial for learning subsequent part shape model and rewriting model. 
To regularize the latent space and avoid high variance, we adopt the KL divergence regularizer towards a standard normal distribution \cite{kingma2013vae,rombach2022stablediffusion}. 
Empirically we find that KL regularizer permits more faithfulness than vector quantization \cite{van2017vqvae,yan2022shapeformer,mittal2022autosdf} for representing continuous shape variations.
Please refer to the supplementary for network and training details.

\subsection{Learning the unified space of part shapes}
\label{sec:unistruct}

Shapes and parts encoded by sequences of patch codes $[\vb{p}_i]$ can still be computationally challenging for rewriting rule learning, as a rule can involve up to dozens of shape parts.
For efficient traversing among shape parts, we embed the parts of different complexities into a \textit{unified} regular latent space $\vb{\Sigma}$, which provides the aligned domain for rewriting network learning (Sec.~\ref{sec:structmapping}).
The embedding is achieved by training $(E_s, D_s)$, which encodes and decodes all parts (including complete shapes) in their normalized scales to focus on geometry.
By ``normalized scales'', we mean that each part is first scaled to fit within a unit cube (with a side length of 2) centered at the origin.
The parts are then instantiated within spatially positioned bounding boxes in the structure rewriting step (Sec.~\ref{subsec:rewrite_network}). 

Noticing that a single code should summarize the patch codes globally and adaptively, we implement $(E_s, D_s)$ as a pair of encoder-decoder transformer \cite{vaswani2017transformer} networks. 
$E_p$ is thus a typical transformer encoder that takes $[\vb{p}_i]$ and a trainable token \texttt{CLS} as input, and produces through self-attention operations the latent codes $[\vb{p}'_i]$ and $\vb{s}\in \vb{\Sigma}$; 
$D_p$ is a parallel transformer decoder that tries to recover $[\vb{p}_i]$ through cross-attention operations.

\begin{figure}[tb]
    \centering
  \includegraphics[width=\linewidth]{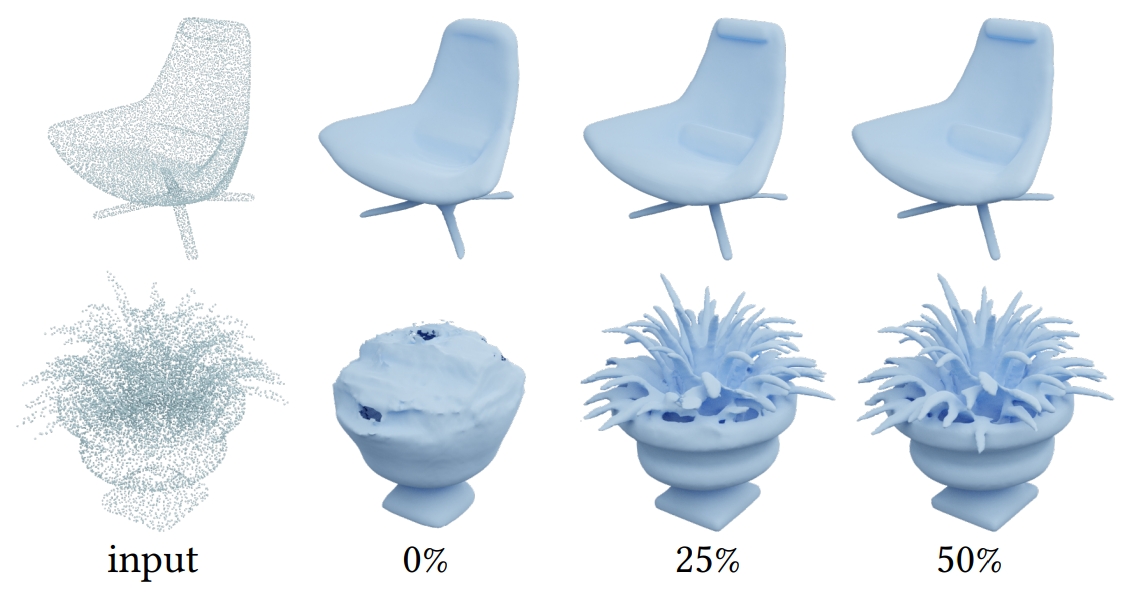}
  \vspace{-12pt}
    \caption{\textbf{Masked input for detailed reconstruction.} 
    With increasing ratios of $[\vb{p}'_i]$ visible to $D_s$, the recovered shapes start with rough overall geometry (0\%) and gradually obtain more fine details (25\%, 50\%). }
    \label{fig:step2_input_condition} \vspace{0mm}
\end{figure}

To encode shapes as diverse as both complete objects and parts of different granularities is a nontrivial task.
We use several tricks to learn a regular encoding:
\begin{itemize}[leftmargin=*]\setlength\itemsep{0mm}
	\item We regularize the latent code space by using a KL-divergence loss from the standard normal distribution, similar to patch latent space learning (Sec.~\ref{sec:svqvae}).
	\item We use parallel decoding \cite{carion2020detr} for the decoder, to minimize the exposure bias \cite{ExposureBias19} of autoregressive transformer decoders that tends to accumulate error along sequence.
	\item To reduce the burden of encoding details from $\vb{s}$ while allowing for the reconstruction of details, we use a masked auto-encoding strategy that sends randomly masked subsequences of $[\vb{p}'_i]$ to cross-attention operations of $D_s$ (Fig.~\ref{fig:pipeline}(b)). The randomized subsequences provide skip-connections for the reconstruction of details, while also ensuring that $\vb{s}$ encodes overall shape structures by not shortcutting it entirely, as shown in Fig.~\ref{fig:step2_input_condition}.
\end{itemize}
Please refer to the supplementary for network and training details.

\section{Rewriting rules learning}
\label{sec:structmapping}

Rewriting rules are hard to specify manually, as they derive from both geometry cues and shape semantics that cannot be described explicitly. 
A data-driven approach is convenient to learn such rules from massive structure transitioning examples, and represents the rules implicitly by using a neural network $R$ to process input shapes and parts and produce output parts organizing the input, which forms the core rewriting model of \name.
Due to the simple syntax of a rewriting model, the neural network $R$ consists mainly of a pair of sequence encoder $E_R$ and decoder $D_R$.
We augment the encoder-decoder with pointer interactions \cite{VinyalsPointer15} to represent hierarchy, use iterative decoding for the decoder to accelerate convergence and enable probabilistic sampling, and synthesize combinatorial training data to enhance the generalization and context-sensitivity of the trained model.

\subsection{Rewriting network}
\label{subsec:rewrite_network}

\paragraph{Part instantiation}
We instantiate a part $s=(\vb{s}, B)$ by representing it with two components: 1) a latent code $\vb{s}\in\vb{\Sigma}$, representing its normalized geometry; 2) a bounding box $B$ with center $\vb{c}\in \mathbb{R}^3$ and $(w,h,l)$ denoting side lengths in three axis directions, representing the actual size and position of the part.

\paragraph{Network input}
The input consists of three groups of data: 1) the patch codes $[\vb{p}_i]$ representing the raw unstructured input geometry, 2) the instantiated parts $[s_i]$ representing a particular decomposition of the shape, and 3) the direction of rewriting $d\in\{\uparrow,\downarrow\}$.
When given no part as input, the network infers leaf node parts corresponding to the shape patches; in this case, $d=\uparrow$ always.
On the other hand, when there are input parts representing a particular structure, the addition of patch codes encoding detailed geometry is shown to improve accuracy (Sec.~\ref{subsec:ablation}).

\paragraph{Encoder} We embed the input patches, parts and directions into latent codes through corresponding feature mapping layers, which are fed into the transformer encoder $E_R$ for contextual feature learning through self-attention layers. We denote the output tokens for the input parts, patches and direction as $[\vb{e}^s_i]$, $[\vb{e}^p_i]$ and $\vb{e}^d$, respectively.

\paragraph{Decoder} The decoder $D_R$ is a parallel decoding transformer network \cite{carion2020detr}, that transforms trainable query vectors $[\vb{q}_i], i{\in}[N_q]$ into output states $[\vb{q}'_i]$, through interleaved self-attention and cross-attention layers that attend to $[\vb{e}^s_i], [\vb{e}^p_i], \vb{e}^d$.
The query set size $N_q$ is chosen to be larger than the number of parts in each shape.
As shown in Fig.~\ref{fig:observation}, the learned queries correspond to consistent structure composition patterns across objects.

\paragraph{Structure prediction.} Given the decoder output $[\vb{q}'_i]$, we apply four projection heads on each token $\vb{q}'_i$ to recover the target parts and hierarchy structure between two levels:
\begin{itemize}\setlength\itemsep{0mm}
    \item The existence head gives the probability $\textrm{P}^{ext}_i \in[0,1]$ of whether the token exists as an output part. 
    \item The geometry head predicts the shape code $\vb{s}'_i\in\vb{\Sigma}$ of the output part.
    \item The bounding box head recovers the bounding box $B'_i$ of the output part.
    \item The pointer head computes for each input part $s_j$, the probability of $q_i$ being adjacent to it by a pointer mechanism \cite{VinyalsPointer15}, to be $\textrm{P}^{adj}_{i,j} = \sigma\ \bigl( MLP_p(\vb{q}'_i)\cdot MLP_p(\vb{e}^s_j) \bigr)$, where $\sigma(\cdot)$ is the sigmoid function, $MLP_p$ is a small 3-layer MLP for feature transformation,
    and the adjacency means that two parts form a child/parent relationship.
\end{itemize}

\paragraph{Matching.} Given the predicted parts that are not aligned with target parts $[\overline{s}_j],j\in[N_{gt}]$ in either numbers or ordering, we first build a matching between the two sets, according to which training losses can be computed.
The matching cost between every pair of predicted part and GT part compares their similarities in all three aspects of existence, geometry and bounding box:
\begin{equation}
      C(q_i,\overline{s}_j) = -w_e\log(\textrm{P}^{ext}_i)  + w_g L_{geo}(\vb{s}'_i, \overline{\vb{s}}_j) + w_b L_{box}(B_i, \overline{B}_j),
      \label{eq:matching} 
\end{equation}
where $L_{geo}(\cdot,\cdot)$ compares the shape codes by squared $l_2$ distance, and $L_{box}(\cdot,\cdot)$ computes the squared $l_2$ distance of two bounding boxes in all dimensions of centers and side lengths.
We use weights $w_e=0.1, w_g=1, w_b=5$ to balance the terms with different scales. Given the matching cost matrix, the linear assignment algorithm \cite{Kuhn1955,Munkres1957} is used to find the optimal matching $M$ that assigns to each predicted part $q_i$ its GT counterpart $M(q_i) \in [\overline{s}_j]\cup\{\phi\}$, where $\phi$ denotes the non-existing counterpart for unmatched parts.

\paragraph{Training loss}

The loss functions measure the difference between matched pairs of predicted and GT parts, in terms of existence, geometry and bounding box, as well as the pointer adjacency with respect to input parts, and the non-existence of unmatched predictions.
Concretely, the first loss term simply sums up the cost terms of matched pairs:
\begin{equation}
	L_{match} = \dfrac{1}{N_{gt}}\sum_{q_i, M(q_i)\neq \phi}{C\bigl(q_i,M(q_i)\bigr)}.
\end{equation}
We find that using the same terms for both matching cost and loss computation stabilizes the training process, which could otherwise fluctuate seriously due to the sudden changes of optimal matching. 
The second loss term measures pointer adjacency:
\begin{equation}
	L_{adj} = \dfrac{1}{N_{gt}\cdot N_{s}}\sum_{q_i, M(q_i)\neq \phi}{\sum_{s_j} BCE\bigl(\textrm{P}^{adj}_{i,j}, \vb{1}\{s_j\sim M(q_i)\}\bigr) },
\end{equation}

where $BCE(\cdot,\cdot)$ computes binary cross entropy, $\vb{1}\{\cdot\}$ is the indicator function, $N_{s}$ is the number of input parts $[s_j]$, and $\sim$ denotes two parts are adjacent in GT hierarchy.
The third loss term encourages the unmatched predictions to exit:
\begin{equation}
	L_{unmatch} = \dfrac{1}{N_q-N_{gt}}\sum_{q_i, M(q_i)=\phi}{-\log(1-\textrm{P}^{ext}_i)}.
\end{equation}
The total loss is the summation of all three terms:
\begin{equation}
	L_{total} = w_m L_{match} + w_a L_{adj} + w_u L_{unmatch}.
	\label{eq:decoder_total_loss}
\end{equation}
We use $w_m=1, w_a=0.1, w_u=0.2$ to balance the loss terms of different scales.

\subsection{Iterative decoding and probabilistic sampling}
\label{subsec:sampling}

\begin{algorithm}[tb]
	\SetAlgoLined 
	\KwIn{training sample $([\vb{p}_i], [s_i], [\overline{s}_i])$} 
	\KwOut{loss function for the training sample}
	
	\tcp{encoder forward}
	$[\vb{e}^p_i], [\vb{e}^s_i] = E_R\left([\vb{p}_i], [s_i]\right)$\;
	
	\tcp{first decoder forward}
	$[\vb{q'}^0_i] = D_R\left([\vb{q}_i]; [\vb{e}^p_i], [\vb{e}^s_i]\right)$\;
	
	\tcp{matching between predictions and GT}
	$M = \textrm{match}\left([\vb{q'}^0_i], [\overline{s}_i]\right)$\;
	
	\tcp{random sampling of queries from matched pairs}
	$Q' = \textrm{random\_sample}\left( [\textrm{P}^{ext}_i|M(q_i)\neq \phi]\right)$\;
	
	\tcp{second decoder forward}
	$[\vb{q'}^1_i] = D_R\left([\vb{q}_i + \vb{1}\{q_i\in Q'\}(\vb{q'}^0_i)]; [\vb{e}^p_i], [\vb{e}^s_i]\right)$\;
	
	\tcp{return loss (Eq.~(\ref{eq:decoder_total_loss})) of both predictions}
	return $\textrm{Loss}\left([\vb{q'}^0_i], [\overline{s}_i]\right) + \textrm{Loss}\left([\vb{q'}^1_i], [\overline{s}_i]\right)$\;
	
	\caption{The training procedure of iterative decoding} 
	\label{alg:iterative_decoding_training}
\end{algorithm}

\begin{algorithm}[tb]
	\SetAlgoLined 
	\KwIn{input test data $([\vb{p}_i], [s_i])$, sampling probability schedule $[\delta_t],t\in [T]$} 
	\KwOut{sampled predictions}
	
	\tcp{encoder forward}
	$[\vb{e}^p_i], [\vb{e}^s_i] = E_R\left([\vb{p}_i], [s_i]\right)$\;
	
	\tcp{initialize the back-substitution tokens}
	$Q = \{\}$, $[\overline{\vb{q}}_i = \vb{0}]$\; 	
	\For{$t \in [T]$}{
		\tcp{decoder forward}
		$[\vb{q'}_i] = D_R\left([\vb{q}_i + \vb{1}\{q_i\in Q\}(\overline{\vb{q}}_i)]; [\vb{e}^p_i], [\vb{e}^s_i]\right)$\;
		\tcp{filter predictions by probability schedule, with random sampling}
		$Q = \textrm{filter\_rand}([\textrm{P}^{ext}_i], \delta_t)$, $[\overline{\vb{q}}_i] = [\vb{q'}_i]$\;
	}
	
	\tcp{return final predictions}
	return $[\overline{\vb{q}}_i]$\;
	
	\caption{The sampling procedure of iterative decoding} 
	\label{alg:iterative_decoding_sampling}
\end{algorithm}

The parallel transformer decoder $D_R$ on one hand avoids the exposure bias of autoregressive decoders that hurts generalization \cite{ExposureBias19}, but on the other hand does not yet support probabilistic sampling of multiple outputs that is essential for structure ambiguity resolution.
Indeed, in the literature this deterministic decoding problem of parallel decoder is known as the multi-modal problem \cite{gu2018nonautoregressive,maskpredict}, saying that for ambiguous input with multiple plausible outputs, the model cannot separate the different targets but instead mixes them indistinctly, which means there can be overlapping and duplicated output components in our case. 
To resolve this issue, \cite{maskpredict,chang2022maskgit} propose an iterative decoding process: given an initial decoded set with conflicts, the iterative decoding process repeatedly removes a subset of lower confident tokens and feeds the rest back to the decoder for a refined output; through this sampling process, one of the many conflicting solutions would be selected and finalized. 
Meanwhile, on a very different note, works using parallel decoding network for detection \cite{li2022DN-detr} have shown that by incorporating partial targets to the decoder queries, the optimal matching of training loss computation stabilizes and the network convergence greatly accelerates.

Based on these observations, we propose to use an iterative decoding scheme for both training and testing, to enable probabilistic sampling of ambiguous solutions and accelerate network training.
In particular, for training, as shown in Alg.~\ref{alg:iterative_decoding_training}, after the first decoder forward pass, we use random sampling of matched predictions based on their existence probability (i.e., $\textrm{P}^{ext}_i > \epsilon_i, \epsilon_i\sim U(0,1)$) to obtain a subset, which is then added to the decoder query set to go for a second pass;
the outputs of these two passes are collected for loss computation.
For test, as shown in Alg.~\ref{alg:iterative_decoding_sampling}, we iterate the decoding process for a specified $T$ cycles following a predefined monotonically decreasing schedule sequence $[\delta_t\in(0,1)], t\in[T]$, and in each iteration $t$ we filter the predicted tokens by $\textrm{P}^{ext}_i > \delta_t + \epsilon_i(1-\delta_t), \epsilon_i\sim U(0,1)$, where $\epsilon$ introduces randomness;
we only add the filtered tokens back to the queries for decoding in the next iteration.
The final predictions are classified as existent or not by their predicted $\textrm{P}^{ext}_i > \gamma$; we have used $\gamma=0.2$ empirically, following previous works \cite{li2022DN-detr}.
Sec.~\ref{subsec:ablation} and \ref{subsec:result_analysis} present experiments demonstrating how the iterative decoding improves structure accuracy significantly and enables sampling of multiple plausible structures.

\subsection{Data augmentation for generalization}
\label{subsec:augmentation}

\begin{algorithm}[tb]
	\SetAlgoLined 
	\KwIn{Training data hierarchy $H$} 
	\KwOut{Augmented training dataset $D$}
	
	\For{$d \in \{\uparrow,\downarrow\}$}{
		\tcp{single shape augmentation}
		\For{shape hierarchy $h\in H$}{
			\For{every two adjacent levels $([s_i],[\overline{s}_i])$ in $h$ with direction $d$}{
				$D = D\cup \{ ([[s_i],[\overline{s}_i]) \}$\;	
				\tcp{randomly mask the two-level hierarchy and form eight new tuples}
				$D = D\cup \textrm{rand\_mask}([s_i],[\overline{s}_i])$\;
			}
		} 
		
		\tcp{paired shape augmentation}
		\For{pairs of shape hierarchies $h_1, h_2\in H$}{
			find eight pairs of adjacent levels from both $h_1, h_2$ with direction $d$\;
			\For{two tuples $([s^1_i],[\overline{s}^1_i])$ in $h_1$, $([s^2_i],[\overline{s}^2_i])$ in $h_2$}{
				$D = D\cup \{ ([s^1_i]\cup[s^2_i],[\overline{s}^1_i]\cup [\overline{s}^2_i] ) \}$\;	
				\tcp{randomly mask the two-level hierarchy and form a new tuple}
				$D = D\cup \textrm{rand\_mask}([s^1_i]\cup[s^2_i],[\overline{s}^1_i]\cup [\overline{s}^2_i])$\;
			}		
		} 
	}
	
	\caption{The data augmentation algorithm} 
	\label{alg:data_augmentation}
\end{algorithm}

The transformer networks $E_R,D_R$ have global contexts of the input patches and parts, which allows for learning context-sensitive rewriting model. 
However, if the training data has limited variations tightly coupled to semantics, it is likely that the rewriting model begins to memorize concurrence of unrelated parts and shapes to derive the outputs, e.g., it may only be able to group four legs into a base given the presence of a chair back, but not able to do so in presence of a table top. 
To avoid such strong coupling to semantics which hurts generalization, and to enable robust reasoning in presence of imperfect data (e.g. missing data due to occlusion in scanning), we design a data augmentation procedure that greatly enhances the model performances as validated in Sec.~\ref{subsec:ablation}.

As shown in Alg.~\ref{alg:data_augmentation}, the data augmentation takes an original dataset (e.g. PartNet~\cite{Mo2019partnet}) with annotated structure hierarchies as input.
It then proceeds to synthesize the input/output pairs of object parts on two neighboring levels in a hierarchy. 
Notably, as shown in line 5, we also use random masking of parts to remove the extra coupling of unrelated parts and to simulate data imperfection.
In particular, we choose from $[s_i], [\overline{s}_i]$ random parts to remove and check: for those parts whose parent node has been removed, they are also removed; for a parent node who misses more than half child nodes, the parent node along with its children are removed.
In this way, the coupling of different subtrees within the hierarchy is broken, which facilitates generalization. 
The missing of partial child nodes also enables the model to recover complete shapes from partial observations (Sec.~\ref{sec:results}).
Note that we can generate a large number of samples for each object by random masking; here we choose eight samples for one object.

Besides single object augmentation, we use paired shape composition to further remove the coupling to predefined categorical semantics.
As shown in Alg.~\ref{alg:data_augmentation} line 8-14, we sample a pair of shape hierarchies $h_1, h_2$, and collect eight arbitrary neighboring levels from the two hierarchies respectively. 
We then merge the two data pairs as one for augmentation, and apply the same random masking of parts as discussed before to enrich the variations. 
Note that there can be overlap between two shapes, which cause ambiguity especially with geometry patch input; we skip shape pairs with intersection over union larger than 10\% to remove such ambiguity.
Again, the combinatorial complexity is so rich that we can generate very large training set (Sec.~\ref{sec:results}).
In practice, we generate roughly the same amounts for single and paired augmentations. 

Finally we note that in the above synthesis process, we always include the patch to leaf part training pairs, which is unique to upward direction rewriting (Sec.~\ref{subsec:rewrite_network}) and enables learning to reconstruct detailed structures from raw unstructured input for general shapes.
\section{Results and Discussion}
\label{sec:results}

\begin{figure*}
    \centering
  \includegraphics[width=\linewidth]{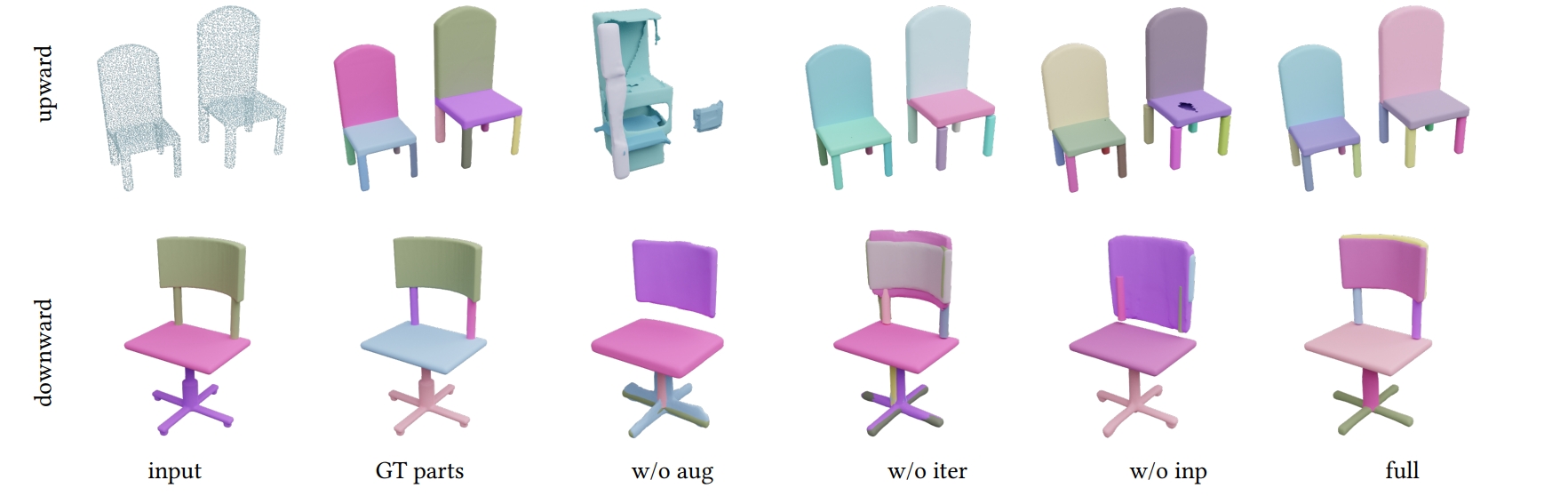}
  \vspace{-16pt}
    \caption{\textbf{Examples of ablation results}. Top row is rewriting in upward direction, bottom row in downward direction. Compared to our full model, the variations on data augmentation, iterative decoding and input geometry conditioning produce less accurate results with missing, duplicated or distorted parts.}
    \label{fig:ablation_mapping}
\end{figure*}

\subsection{Implementation details}
\label{subsec:implementation}

We mainly use PartNet \cite{Mo2019partnet} hierarchies as training data. Following StructureNet \cite{mo2019structurenet}, we use the same training/validation/test splits for training the three modules, where the split ratios are 7:1:2.
In addition, to facilitate testing the extension of our model, we include the rest $34$ categories from ShapeNet \cite{chang2015shapenet} that are not covered by PartNet.
For this extra data without labeling, we use heuristics such as groups in obj files and connected components to obtain leaf node parts. 
While such heuristic labeling is generally of low quality and inconsistent, it potentially allows the network to recognize parts not seen in PartNet categories.
We center each shape to origin and normalize it by uniform scaling to fit into $[-1,1]^3$.

To train the latent geometry auto-encoding network $(E_p,D_p)$, we use complete object shapes from both ShapeNet and PartNet.
We use augmentations of random rotation, scaling, and translations to enrich the diversity of training data.

To train the part shape auto-encoding network $(E_s,D_s)$, we use complete shapes from PartNet and ShapeNet, as well as parts of all levels, for training.
We also apply data augmentation that rotates parts around axes for multiples of $90^\circ$ to enrich shape diversity and be robust to rotations.

Finally, we use PartNet hierarchies and ShapeNet leaf nodes to train the rewriting network $(E_R,D_R)$. 
In total, we have generated $9.03 M$ training samples.
Among them, $8.61M$ are from PartNet and the rest $0.42M$ from ShapeNet.
For the $18K$ PartNet training shapes, we first non-uniformly scale each object for 8 times to increase diversity, and then follow the augmentation procedure (Sec.~\ref{subsec:augmentation}, Alg.~\ref{alg:data_augmentation}) to generate the combinatorial dataset. 

We trained the model on 8 NVIDIA V100 GPUs, each with 24GB memory. 
Please refer to Sec. 4 in the supplemental for more detailed analysis about the training and inference costs of each component.

\begin{table}[t]
\fontsize{8}{12}\selectfont
  \centering
  \caption{\textbf{Ablation results} on 1k random samples from test set. 
  Iter, Aug and Inp stand for iterative decoding, combinatorial data augmentation and input patch conditioning, respectively.
  Exist measures the existence of predicted parts against GT parts, Ptr the predicted pointers between input and output parts against GT pointers, and Box and Code the $l_1$ errors between predictions and GT over bounding box and code, respectively.
  }
    \begin{tabular}{cccccccc}
    \toprule[1pt]        
    \textbf{Config} & \textbf{Iter} & \textbf{Aug}  & \textbf{Inp}  &  \textbf{Exist} $\uparrow$ & \textbf{Ptr} $\uparrow$ & \textbf{Box} $\downarrow$ & \textbf{Code} $\downarrow$\\
    \midrule
    w/o aug  & \checkmark &     &   \checkmark      & 0.83 & 0.95  & 0.04 & 0.60 \\
    w/o iter &  &  \checkmark  &   \checkmark       & 0.74 & \textbf{0.98}  & \textbf{0.03} & 0.56 \\
    w/o inp  & \checkmark &  \checkmark      &      & 0.83 & \textbf{0.98}  & \textbf{0.03} & 0.57 \\
    full     & \checkmark & \checkmark & \checkmark&  \textbf{0.84} & \textbf{0.98}  & \textbf{0.03} & \textbf{0.55} \\
    \bottomrule[1pt]
    \end{tabular}%
  \label{tab:ablation}%
  \vspace{0mm}
\end{table}%

\subsection{Ablation study}
\label{subsec:ablation}

We conduct ablation tests on major components of our rewriting network $(E_R,D_R)$.
Further ablation tests on part shape representation learning can be found in the supplemental.

\paragraph{Setup and metrics}

The ablation settings are evaluated on 1000 samples randomly selected from the test set, covering all shape categories, upward/downward directions, and levels within hierarchies.
We evaluate metrics corresponding to the training loss terms (Sec.~\ref{subsec:rewrite_network}).
In particular, let $[\overline{s}_i]$ be GT output parts, $[q_i]$ the predicted parts, and $M(q_i)$ their matching, we compute:
\begin{itemize}
    \item the existence (\textbf{Exist}) that measures the F-score of predicted part existence. 
    The true positive correspondence of a predicted part $q_i$ ($\textrm{P}^{ext}_i>0.5$) to GT is established iff $M(q_i)\neq \phi$.
    
    \item the pointer accuracy (\textbf{Ptr}), as measured by the F-score of predicted pointer pairs. For a GT pair $\overline{\textrm{P}}^{adj}_{j,k}$ and a predicted pair $\textrm{P}^{adj}_{i,k} > 0.5$, the correspondence is established iff $M(q_i)=s_j$.
    
    \item bounding box accuracy (\textbf{Box}) and shape code accuracy (\textbf{Code}), both computed as the $l_1$-norm of difference vectors from GT boxes and codes (c.f. Eq.~(\ref{eq:matching})).
\end{itemize}

\paragraph{Combinatorial data augmentation}
To assess the importance of random combination of structures (Sec.~\ref{subsec:augmentation}) for learning local rewriting rules that generalize to different shapes, we exclude the combinatorial composition from the data augmentation pipeline.
As shown in Tab.~\ref{tab:ablation}, the trained network degrades in terms of both geometric fidelity (box and code accuracy) and pointer accuracy.
Visually in Fig.~\ref{fig:ablation_mapping} it fails seriously on the input containing two separate chairs, and tends to miss important features, confirming its reduced robustness and generalization.

\paragraph{Iterative decoding for sampling and disambiguation}
Iterative decoding enables probabilistic sampling and the accommodation of ambiguous structures, which is vital for cross-category training and generalization.
Thus, it can be expected that by switching to a fixed decoding process for both training and testing, the structure accuracy is impacted the most.
This is confirmed by the largely reduced existence accuracy in Tab.~\ref{tab:ablation}.
Visually as shown in Fig.~\ref{fig:ablation_mapping}, without iterative decoding, the network may produce duplicated parts, demonstrating its confusion of multiple solutions.

\begin{figure}[t]
\centering

  \includegraphics[width=\linewidth]{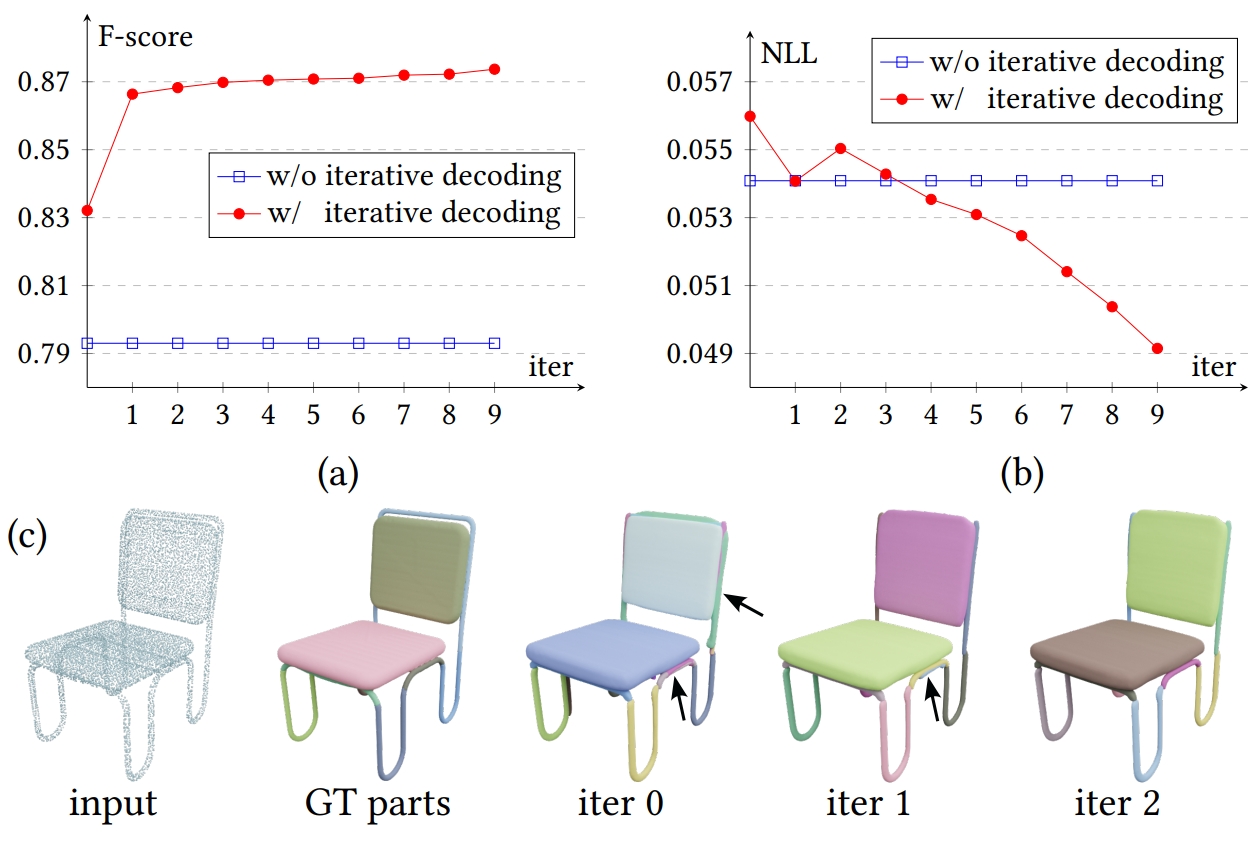}
  \vspace{-16pt}

    \caption{\textbf{Iterative decoding of $D_R$}. \textbf{(a) Accuracy}: The accuracy of predictions measured by F-score increases along iterations.  \textbf{(b) NLL}: The probability of GT output measured by negative log-likelihood increases along iterations. \textbf{(c) Visualization}: Given the point cloud input, the iterations first obtain covering parts and then gradually remove redundant ones (highlighted by arrows).
    }
    \label{fig:ablation_iter_decoding}
    \vspace{0mm}
\end{figure}

To further analyze the iterative decoding process, we closely inspect the intermediate results between iterations.
In particular, we show the existence F-score of predicted parts against GT and the negative log-likelihood (NLL) of GT parts when they are substituted for matched predictions.
As shown in Fig.~\ref{fig:ablation_iter_decoding}, as the iteration goes, the F-score increases steadily and the NLL decreases significantly, both surpassing the baseline without iterative decoding quickly;
visually, the iterations quickly remove redundant part predictions.
Based on these observations, we find that 5 iterations are usually sufficient for obtaining a good solution, and have reported other results using 5 iterative decoding passes.

\paragraph{Input geometry conditioning for rewriting}
Patch codes $[\vb{p}_i]$ encoding input geometry can be omitted when the input parts $[s_i]$ are available (Sec.~\ref{subsec:rewrite_network}).
However, we note that having the input patches enhances rewriting quality.
As shown in Tab.~\ref{tab:ablation}, without input patches the produced structures tend to have larger geometric errors reflected in code distance; Fig.~\ref{fig:ablation_mapping} shows that indeed without input patches, the network produces less accurate shapes.  

To further validate the effectiveness of each module,
we conduct an additional ablation study on the entire chair category, with reconstruction performance metrics and more visual examples. Please refer to Table 2 and Fig. 2 in the supplemental for more details.

\subsection{Comparison on structured reconstruction}
\label{subsec:comparison_recon}
In this section we evaluate the performance of reconstruction from point clouds that are either complete or incomplete due to scanning. 

\paragraph{Baselines} 
StructureNet~\cite{mo2019structurenet} is the baseline method for structured shape reconstruction from point cloud.
It first learns the implicit latent space of category-specific hierarchies, and then builds a PointNet++ encoder that maps point cloud input to the corresponding latent code for structured reconstruction.
In comparison, we use rewriting to obtain structures from unstructured input point cloud and hierarchies through its iterative applications, and thus is not limited by category-specific latent space training.

\begin{table}[t]
\fontsize{8}{12}\selectfont
  \centering
  \caption{ \textbf{Comparison on shape reconstruction}. StructureNet \cite{mo2019structurenet} trains specific models for each object category, while AR baseline and our model are evaluated across categories. 
  }
    \begin{tabular}{ccrrr}
    \toprule[1pt]
    \multirow{2}[4]{*}{\textbf{Category}} & \multirow{2}[4]{*}{\textbf{Method}} & \multicolumn{2}{c}{\textbf{Geometry}} & \textbf{Structure}  \\
    \cmidrule{3-5}          &       & \textbf{CD}{\scriptsize$\times 10^{-2}$}$\downarrow$ & \textbf{F1}$\uparrow$ & \textbf{Hier} $\downarrow$\\
    
    \midrule
    \midrule
    \multirow{2}[2]{*}{Chair}  & StructureNet	  & 1.36	  & 0.71	  & 0.47 \\
    & AR        &    0.68  &  0.86  &  0.59 \\
    & Ours	  & \textbf{0.37}	  & \textbf{0.95}	  & \textbf{0.43} \\

    \midrule
    \multirow{2}[2]{*}{Table} & StructureNet	  & 3.50  	  & 0.49	  & 0.67 \\
    & AR & 0.62 & 0.89 & 0.72\\
    & Ours	  & \textbf{0.37}	  & \textbf{0.95}	  & \textbf{0.55} \\ 

    \midrule
    \multirow{2}[2]{*}{Cabinet} & StructureNet	  & 1.81	  & 0.64	  & 0.72 \\
    & AR & 0.75 & 0.85 & 0.87 \\
    & Ours	  & \textbf{0.51}	  & \textbf{0.92}	  & \textbf{0.70} \\

    \bottomrule[1pt]
    \end{tabular}%
  \label{tab:reconeval}%
  \vspace{0mm}
\end{table}%
\begin{figure}
    \centering
  \includegraphics[width=\linewidth]{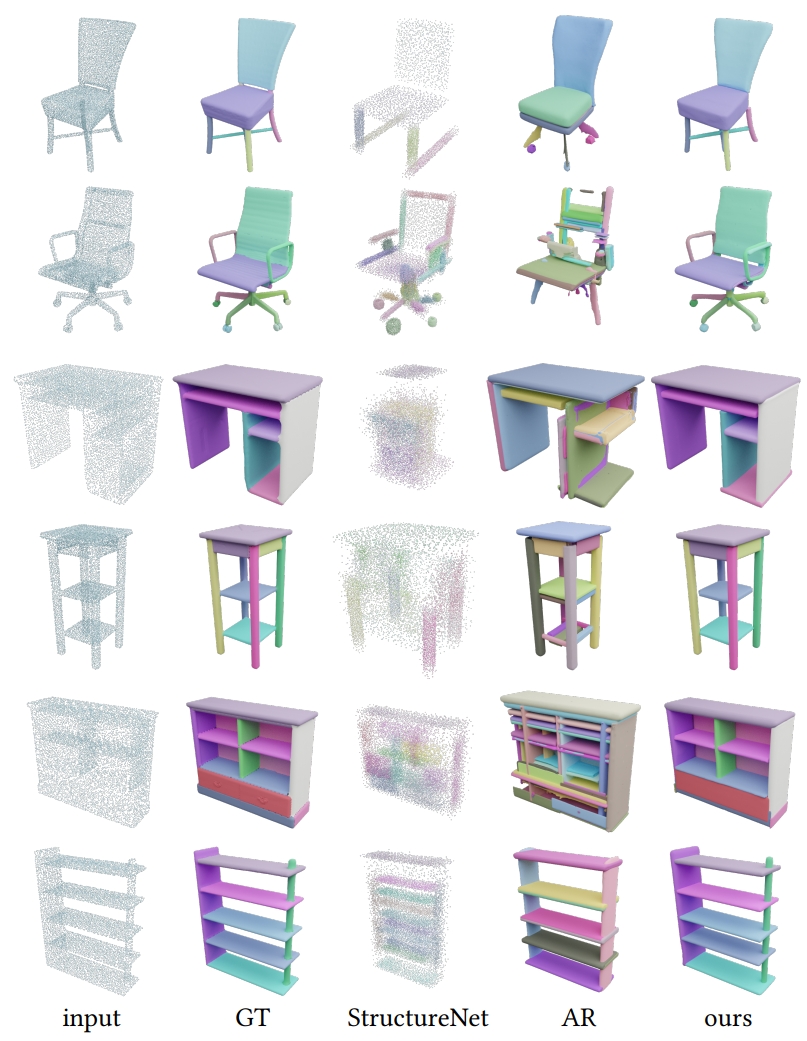}
  \vspace{-12pt}
    \caption{\textbf{Structured shape reconstructions}. StructureNet and AR baseline learn latent representations from which the complete structures and detailed geometry are expanded in one pass. In contrast, our approach recover the leaf parts from input points by local rewriting. 
    }
    \label{fig:comp}\vspace{0mm}
\end{figure}

To verify the effectiveness of the proposed local rewriting scheme, we design a transformer-based autoregressive (AR) baseline that outputs the whole shape structure in one pass, instead of multiple rewriting steps. 
Specifically, given a hierarchically structured shape, we serialize the hierarchy into a sequence by bread-first traversal, where each part node is encoded as the tuple $(\vb{s}, B, idx)$ where $idx \in \vb{N}$ specifies the parental node index in the sequence.
For fair comparison, the AR baseline adopts a transformer encoder-decoder of the same scale as $(E_R,D_R)$, except that the encoder of AR baseline consumes geometry patches only, and the decoder recovers the serialized hierarchy in one pass.
We train the AR baseline on the same data as ours (Sec.~\ref{subsec:augmentation}), but remove the combinatorial part augmentation which produces unrooted hierarchies not representable by the AR baseline.

To further validate the efficiency of structured reconstruction by rewriting, we also compare with ShapeFormer \cite{yan2022shapeformer} which is a state-of-the-art shape completion model that models shapes as patch sequences through a powerful autoregressive transformer, which shares much similarity with our patch based shape representation and transformer network backbone and differs mainly in the unstructured shape representation.

We compare with StructureNet on both complete and incomplete point clouds, with the AR baseline on complete point clouds, and with ShapeFormer on incomplete virtual scans that it is designed for, although our network has never been trained on incomplete data.
The complete point clouds are uniformly sampled from input shapes for 10k points each.
To simulate virtual scans, we sample $3\sim 5$ viewpoints randomly and uniformly on the bounding sphere of a shape, and fuse the partial scans into an incomplete point cloud; normal vectors as additional input features for StructureNet and our model are estimated by local PCA analysis.

\paragraph{Metrics}

We evaluate the results in terms of structural accuracy and geometric fidelity.
To measure hierarchical structure accuracy (\textbf{Hier}), we follow \cite{mo2019structurenet,mo2020pt2pc} to compute the tree node correspondences based on geometric proximity, and then compute the tree editing distance \cite{PaassenUTED21} over GT hierarchy.
To measure geometric fidelity, we follow previous works \cite{mo2019structurenet,Yang2022DSGNet} and compute surface chamfer distance (\textbf{CD}) and F-score (\textbf{F1}) from GT surfaces for merged parts.

\paragraph{Reconstruction from complete point cloud}

In terms of geometric fidelity, as shown in Tab.~\ref{tab:reconeval} and Fig.~\ref{fig:comp}, our leaf node reconstructions are more accurate than StructureNet or AR baseline.
Our improved accuracy can be attributed to two factors: first, our rewriting model directly outputs possible leaf node structures from raw input, without going through a single latent code as StructureNet does, or reproducing a complete hierarchy as AR baseline does which is likely to accumulate errors; second, we obtain increased robustness when the local rewriting model can be trained on significantly larger and diverse dataset across categories and combinations (c.f. Sec.~\ref{subsec:ablation}).

In terms of structure accuracy, as shown in Tab.~\ref{tab:reconeval}, our results are slightly better than StructureNet and significantly better than the AR baseline, although we do not train specific models for each category and thus cannot exploit categorical priors that StructureNet relies on for resolving structural ambiguity. 
Such generalizable performance across categories can be attributed to our local and probabilistic rewriting learning for accommodating ambiguous structures.

The AR baseline clearly shows degraded performances on both geometry and structure (Tab.~\ref{tab:reconeval}, Fig.~\ref{fig:comp}), although its geometry improves over StructureNet due to the high-resolution shape representation (Sec.~\ref{sec:shape_rep_learning}).
The inferior performance of AR baseline can be attributed to the limited generalization of auto-regressive transformer against compositional complexity \cite{LimitsTransformer23}, compounded only by its lack of training on combinatorial augmentation due to its one-pass hierarchy generation.
In contrast, our local rewriting approach robustly handles such tasks, in a similar spirit to the chain-of-thought multiple step reasoning for language models \cite{CoT22,ToT23} (c.f. Fig.~\ref{fig:sampling}).

\begin{table}[t]
\fontsize{8}{12}\selectfont
  \centering
  \caption{ \textbf{Comparison on shape reconstruction from partial scans}. StructureNet \cite{mo2019structurenet}  trains specific models for each object category, while ShapeFormer \cite{yan2022shapeformer} and our model are evaluated across categories. Only ShapeFormer has been trained on partial data,
  while StructureNet and our method are not trained for this task}.
    \begin{tabular}{ccrrr}
    \toprule[1pt]
    \multirow{2}[4]{*}{\textbf{Category}} & \multirow{2}[4]{*}{\textbf{Method}} & \multicolumn{2}{c}{\textbf{Geometry}} & \textbf{Structure}  \\
    \cmidrule{3-5}          &       & \textbf{CD}{\scriptsize$\times 10^{-2}$}$\downarrow$ & \textbf{F1}$\uparrow$ & \textbf{Hier} $\downarrow$\\
    
\midrule
\multirow{3}[2]{*}{Chair}  & ShapeFormer & 0.60 &	0.90	& - \\
& StructureNet  &1.37 & 	0.71	& 0.47 \\
& {Ours}  & \textbf{0.37}	& \textbf{0.94}	& \textbf{0.43}\\

\midrule
\multirow{3}[2]{*}{Table}  & ShapeFormer & 0.73	& 0.90 & 	-\\
& StructureNet  &3.44 &	0.49 &	0.65\\
& {Ours}  & \textbf{0.38}	 & \textbf{0.95} &	\textbf{0.55}\\

\midrule
\multirow{3}[2]{*}{Cabinet}  & ShapeFormer & 0.62 &	\textbf{0.90} &	-\\
& StructureNet  &1.94	& 0.62  &	\textbf{0.74}\\
& {Ours}  & \textbf{0.61}	& 0.89	&   0.75\\

    \bottomrule[1pt]
    \end{tabular}%
  \label{tab:eval_shape_comp}%
\end{table}%
\begin{figure}
    \centering
  \includegraphics[width=\linewidth]{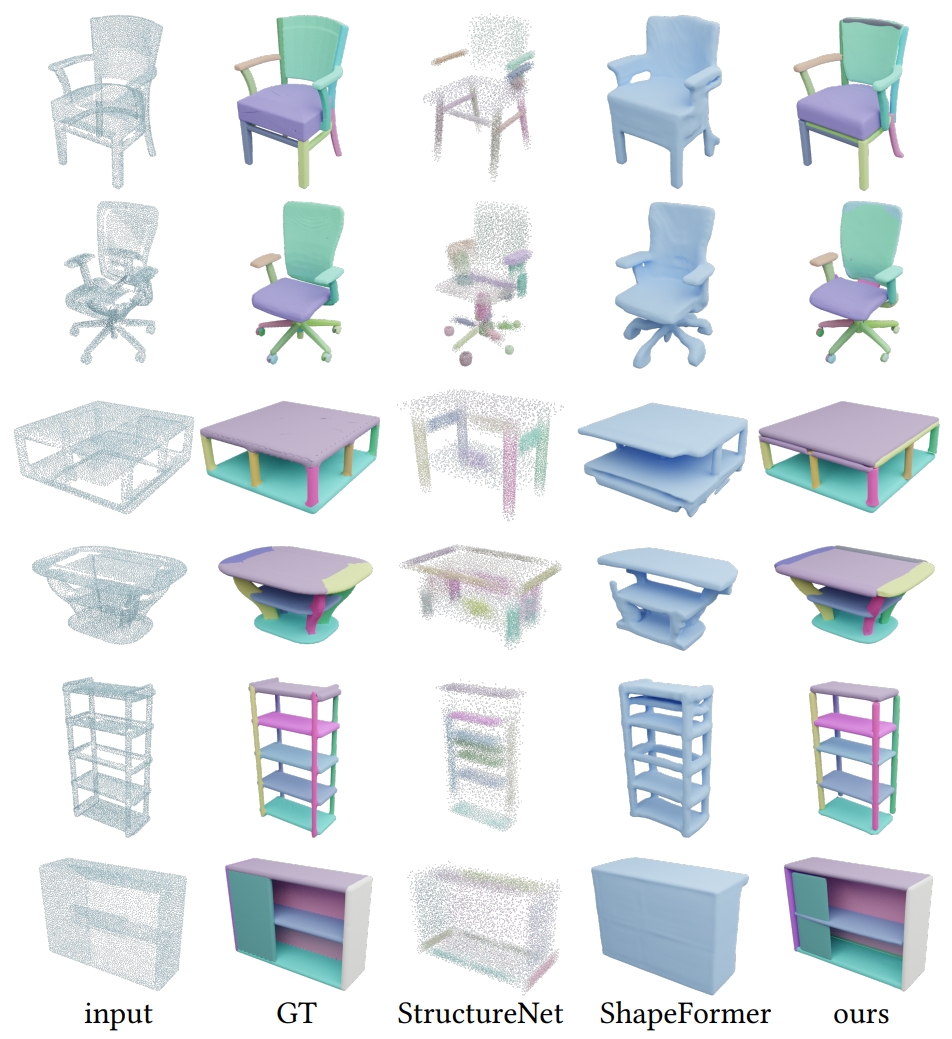}
  \vspace{-12pt}
    
    \caption{\textbf{Reconstructions from partial scans}. StructureNet has regular reconstructions that nevertheless do not capture the input well. ShapeFormer proceeds unstructured and recovers less regular shapes. Our results have both structure regularity and faithfulness.}
    \label{fig:vis_completion}
\end{figure}

\paragraph{Reconstruction from partial scans}

When facing partial point clouds, the reconstruction gets more challenging as shape completion should be done simultaneously.
In this case, methods that are structure aware have the potential of producing more regular shapes due to the regularity of shape parts.
Indeed, when we compare our results with ShapeFormer (Fig.~\ref{fig:vis_completion}, Tab.~\ref{tab:eval_shape_comp}), a state-of-the-art transformer based autoregressive model for shape completion, we find that our results are generally more faithful to the input partial scans, while completing the missing regions with great regularity.
When compared to StructureNet, although its results maintain the regularity of structure awareness, our results again show more accurate structures and detailed surfaces (Fig.~\ref{fig:vis_completion}, Tab.~\ref{tab:eval_shape_comp}), as in the case of reconstruction from complete point clouds.

\subsection{Comparison on structured generation}
\label{subsec:comparison_gen}

In this section, we evaluate the generative modeling performance enabled by \name. 
Since our model is trained across categories and all possible shapes and parts are encoded into a unified latent space $\vb{\Sigma}$, 
it is hard to directly perform unconditional sampling for generation evaluation, in contrast to existing methods \cite{mo2019structurenet,RobertsLSD2021} that train category-specific VAE models for generation.
Thus, for comparison, we present two sampling strategy of latent vectors to perform generation with different conditions:
1) category-conditioned generation, namely the unconditional generation in \cite{mo2019structurenet,RobertsLSD2021} (Sec. \ref{sec:gen_cat-condition});
2) shape-conditioned generation (Sec. \ref{sec:gen_shape_cond}).

\subsubsection{Category-conditioned generation}\label{sec:gen_cat-condition}

\begin{figure*}
    \centering
  \includegraphics[width=\linewidth]{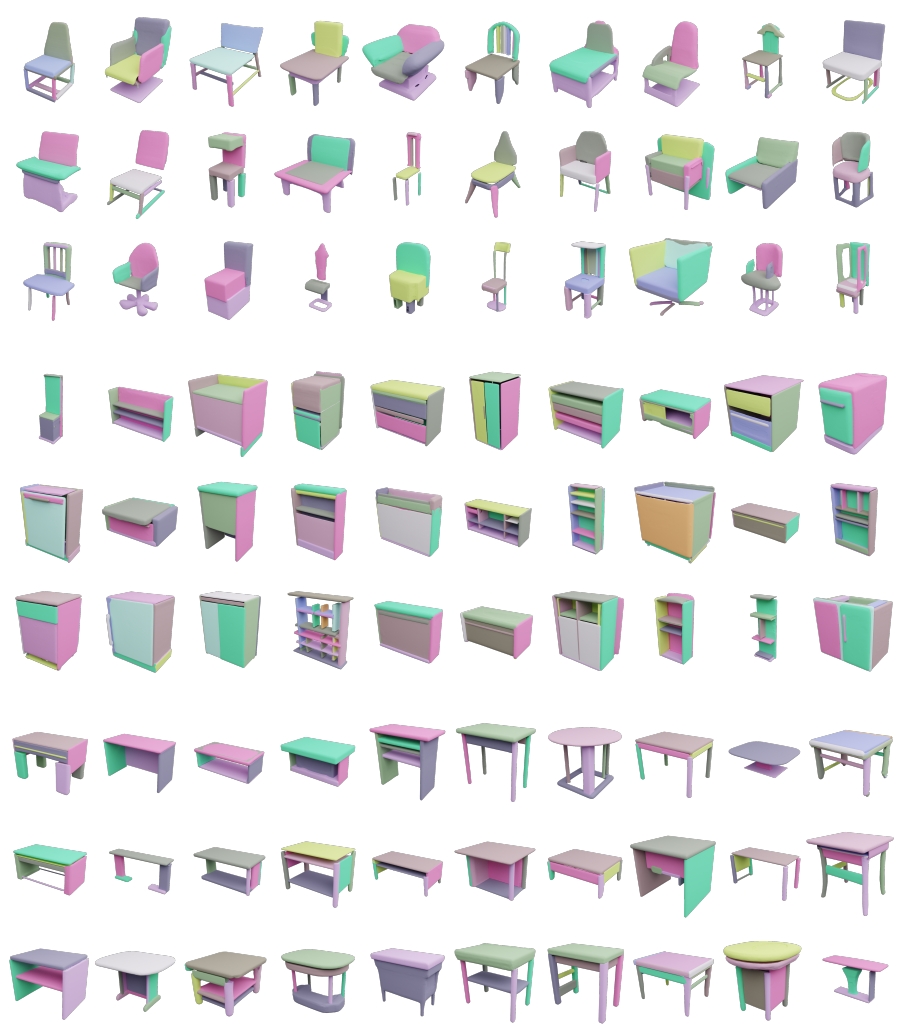}
  \vspace{-12pt}

    \caption{
    \textbf{Gallery of category-conditioned generation.} We present generation results for three categories: Chair, Cabinet and Table. For each category, we fit a multivariate Gaussian distribution and randomly sample latent vectors from this distribution. The sampled latent vectors are then decoded into diverse, structured shapes.
    }
    \label{fig:uncond_gen}
\end{figure*}

To enable our method with a diverse generation capability, 
we present a simple sampling strategy to support unconditional or category-conditioned sampling. We achieve this by fitting a multivariate Gaussian distribution for each category based on the shape codes from the training set. By sampling this distribution, we can perform unconditional generation of novel shapes. 
We have compared these results against StructureNet, and find that our results outperform StructureNet results significantly in FPD and Coverage (Table \ref{tab:eval_shape_gen}). 

Specifically, for a given shape category, we first encode all the complete shapes in the training set into shape codes $[\vb{z}_i]$ using our patch encoder \( E_p \) and part shape encoder \( E_s \). From this set of shape codes, we fit a multivariate Gaussian distribution $\mathcal{N}(\mu,\Sigma)$ by calculating the mean and covariance:
\[
\mu = \frac{1}{N} \sum_{i=1}^{N} \vb{z}_i,\quad 
\Sigma = \frac{1}{N-1} \sum_{i=1}^{N} (\vb{z}_i - \mu)(\vb{z}_i - \mu)^T
\]
where \( N \) is the number of shapes in the category, and \( \mu, \Sigma \) are the mean code and covariance matrix of the shape codes, respectively.

Using this distribution, we can perform unconditional sampling by drawing samples $z {\sim} \mathcal{N}(\mu, \Sigma)$ from the learned multivariate Gaussian distribution. These shape codes can then be decoded into meaningful shapes using the generation mechanism of our model.

We randomly sampled \(1,000\) latent vectors from the distribution and compared our generation results with those from StructureNet and our shape-conditioned sampling strategy, as shown in Table \ref{tab:eval_shape_gen}. 
Our results outperform StructureNet, particularly in terms of the FPD metric. Fig. \ref{fig:uncond_gen} also provides qualitative visualizations of the generated shapes, demonstrating that our approach achieves superior performance in both quality and diversity.

\begin{table}[t]
\fontsize{8}{12}\selectfont
  \centering
  \caption{ \textbf{Comparison on structured shape generation}. StructureNet \cite{mo2019structurenet}  trains specific models for each object category, while our model is evaluated across categories. 
  Ours-category and Ours-shape indicate category-conditioned and shape-conditioned generation, respectively.
  }
    \begin{tabular}{ccrrr}
    \toprule[1pt]
    \textbf{Category} & \textbf{Method} & \textbf{FPD}$\downarrow$ & \textbf{COV}$\downarrow$ & \textbf{MMD} $\downarrow$\\
    
    \midrule
    \multirow{3}[2]{*}{Chair}  & StructureNet	  &  4.67	  &0.89& \textbf{0.58} \\
& Ours-category & 2.86 & 0.82 & 0.76 \\
& Ours-shape	   & \textbf{2.63}  & \textbf{0.70} & 0.65 \\ 

    \midrule
    \multirow{3}[2]{*}{Table} & StructureNet	  &  6.07	  &  1.43	  & 0.55 \\
& Ours-category	   & 2.86 & 0.82 & 0.76 \\
& Ours-shape	 &  \textbf{1.98} & \textbf{0.66} & \textbf{0.53} \\

    \midrule
    \multirow{3}[2]{*}{Cabinet} & StructureNet	  &  8.46	  &  	1.15  &  \textbf{0.67}\\
& Ours-category	   & 4.7 & 0.56  &  0.76 \\
& Ours-shape & 	\textbf{3.88}  &  \textbf{0.58} & 0.71  \\

    \bottomrule[1pt]
    \end{tabular}%
  \label{tab:eval_shape_gen}%
  \vspace{0mm}
\end{table}%

\begin{figure}
    \centering
  \includegraphics[width=\linewidth]{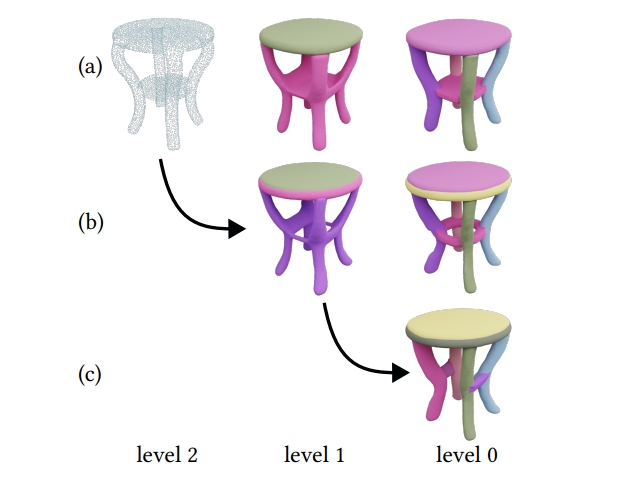}
  \vspace{-24pt}
    
    \caption{\textbf{Level-of-detail generation}. (a): downward rewriting without shape code perturbation. (b): perturbation on the root node only produces a very different structure from (a). (c): following (b), perturbation further on the level 1 table base node changes its detailed structure from (b). }
    \vspace{-12pt}
    \label{fig:vis_gen_lod}
\end{figure}

\begin{figure}[t]
\centering

{\scalefont{1.4}
\begin{tikzpicture}[scale=0.5]
\begin{axis}[
    unit vector ratio=150 175 1,
    xlabel={FPD},
    ylabel={Geo div},
    axis line style={->},
    axis lines=middle,
    xmin=2.3, xmax=5.1,
    ymin=0.2, ymax=1.8,
    xtick={2.0, 2.5, 3.0, 3.5, 4.0, 4.5, 5.0},
    ytick={0.2,0.4,0.6,0.8,1.0,1.2,1.4, 1.6},
    legend style={at={(0.8,0.4)}},
    ymajorgrids=true,
    grid style=dashed,
]
    \addplot+[
        color=red,
        mark=square,
        ]
    coordinates {
        (2.57, 0.26)
        (2.55, 0.34)
        (2.63, 0.46)
        (2.74, 0.68)
        (3.05, 1.09)
        (3.69, 1.61)
    };
    
    \addplot[dashed, color=blue] 
        coordinates {
            (4.67,0)
            (4.67,2)
        };
    \legend{Ours,\; SNet}
\end{axis}
\end{tikzpicture}
}
\hfill
{\scalefont{1.4}
\begin{tikzpicture}[scale=0.5]
\begin{axis}[
    unit vector ratio=150 350 1,
    xlabel={FPD},
    ylabel={Structure div},
    axis line style={->},
    axis lines=middle,
    xmin=2.3, xmax=5.1,
    ymin=0.3, ymax=1.1,
    xtick={2.0, 2.5, 3.0, 3.5, 4.0, 4.5, 5.0},
    ytick={0.3,0.4,0.5, 0.6, 0.7, 0.8, 0.9, 1.0},
    legend style={at={(0.8,0.4)}},
    ymajorgrids=true,
    grid style=dashed,
]
    \addplot[color=red,mark=square,]
        coordinates {
            (2.57, 0.43)
            (2.55, 0.54)
            (2.63, 0.64)
            (2.74, 0.72)
            (3.05, 0.81)
            (3.69, 0.89)
        };
    \addplot[dashed, color=blue] 
        coordinates {
            (4.67,0)
            (4.67,2)
        };
    \legend{Ours,\; SNet}
    
\end{axis}
\end{tikzpicture}
}
\\ {\vspace{-2mm} \small (a) chair} \\
{\scalefont{1.4}
\begin{tikzpicture}[scale=0.5]
\begin{axis}[
    unit vector ratio=150 325 1,
    xlabel={FPD},
    ylabel={Geo div},
    axis line style={->},
    axis lines=middle,
    xmin=1.5, xmax=6.7,
    ymin=0.2, ymax=1.8,
    xtick={1.5, 2.0, 2.5, 3,0, 3.5,4.0, 5.0, 6.0},
    ytick={0.2,0.4,0.6,0.8,1.0,1.2,1.4, 1.6},
    legend style={at={(0.8,0.4)}},
    ymajorgrids=true,
    grid style=dashed,
]
\addplot[
    color=red,
    mark=square,
    ]
    coordinates {
        (1.89, 0.34)
        (1.92, 0.38)
        (1.98, 0.47)
        (2.13, 0.71)
        (2.68, 1.16)
        (3.42, 1.71)
    };
    \addplot[dashed, color=blue] 
        coordinates {
            (6.07,0)
            (6.07,2)
        };
    \legend{Ours,\; SNet}
\end{axis}
\end{tikzpicture}
}
\hfill
{\scalefont{1.4}
\begin{tikzpicture}[scale=0.5]
\begin{axis}[
    unit vector ratio=15 130 1,
    xlabel={FPD},
    ylabel={Structure div},
    axis line style={->},
    axis lines=middle,
    xmin=1.5, xmax=6.7,
    ymin=0.49, ymax=0.91,
    xtick={1.5, 2.0, 2.5, 3,0, 3.5, 4.0, 5.0, 6.0},
    ytick={0.55, 0.6, 0.65, 0.7, 0.75, 0.8, 0.85},
    legend style={at={(0.8,0.4)}},
    ymajorgrids=true,
    grid style=dashed,
]
    \addplot[color=red,mark=square,]
        coordinates {
            (1.89, 0.54)
            (1.92, 0.58)
            (1.98, 0.62)
            (2.13, 0.70)
            (2.68, 0.77)
            (3.42, 0.83)
        };
    \addplot[dashed, color=blue] 
        coordinates {
            (6.07,0)
            (6.07,2)
        };
    \legend{Ours,\; SNet}
\end{axis}
\end{tikzpicture}
}
\\ {\vspace{-2mm} \small (b) table} \\
{\scalefont{1.4}
\begin{tikzpicture}[scale=0.5]
\begin{axis}[
    unit vector ratio=150 333 1,
    xlabel={FPD},
    ylabel={Geo div},
    axis line style={->},
    axis lines=middle,
    xmin=3.5, xmax=8.5,
    ymin=0.3, ymax=1.8,
    xticklabels={3.5, 4.0, 4.5, 5.0, 5.5, 6, 7, 8},
    ytick={0.2,0.4,0.6,0.8,1.0,1.2,1.4, 1.6},
    legend style={at={(0.8,0.4)}},
    ymajorgrids=true,
    grid style=dashed,
]
\addplot[
    color=red,
    mark=square,
    ]
    coordinates {
        (3.85,0.43)
        (3.87,0.50)
        (3.88,0.58)
        (4.15,0.78)
        (4.80,1.12)
        (5.63,1.61)
    };

    \addplot[dashed, color=blue] 
        coordinates {
            (8.46,0)
            (8.46,2)
        };
    \legend{Ours,\; SNet}
\end{axis}
\end{tikzpicture}
}
\hfill
{\scalefont{1.4}
\begin{tikzpicture}[scale=0.5]
\begin{axis}[
    unit vector ratio=150 625 1,
    xlabel={FPD},
    ylabel={Structure div},
    axis line style={->},
    axis lines=middle,
    xmin=3.5, xmax=8.5,
    ymin=0.3, ymax=1.1,
    xticklabels={3.5, 4.0, 4.5, 5.0, 5.5, 6, 7, 8},
    ytick={0.3,0.4,0.5, 0.6, 0.7, 0.8, 0.9, 1.0},
    legend style={at={(0.8,0.4)}},
    ymajorgrids=true,
    grid style=dashed,
]
    \addplot[color=red,mark=square,]
        coordinates {
            (3.85,0.56)
            (3.87,0.65)
            (3.88,0.71)
            (4.15,0.77)
            (4.80,0.82)
            (5.63,0.85)
        };
    \addplot[dashed, color=blue] 
        coordinates {
            (8.46,0)
            (8.46,2)
        };
    \legend{Ours,\; SNet}
    
\end{axis}
\end{tikzpicture}
}
\\ {\vspace{-2mm} \small (c) cabinet}
    \vspace{-3mm}
    \caption{\textbf{Generation diversity} with respect to different perturbation strengths. \textbf{Left}: geometric diversity.  \textbf{Right}:  structure diversity. 
    In each figure, the increasing noise level leads to increased diversity but also decreased similarity to test data (shown by FPD). Nonetheless, the results have better fidelity than StructureNet results. We use noise strength $\sigma=[0, 0.2, 0.4, 0.6, 0.8, 1.0]$ for the six points in each plot.
    }
    \label{fig:eval_gen_div}
    \vspace{0mm}
\end{figure}
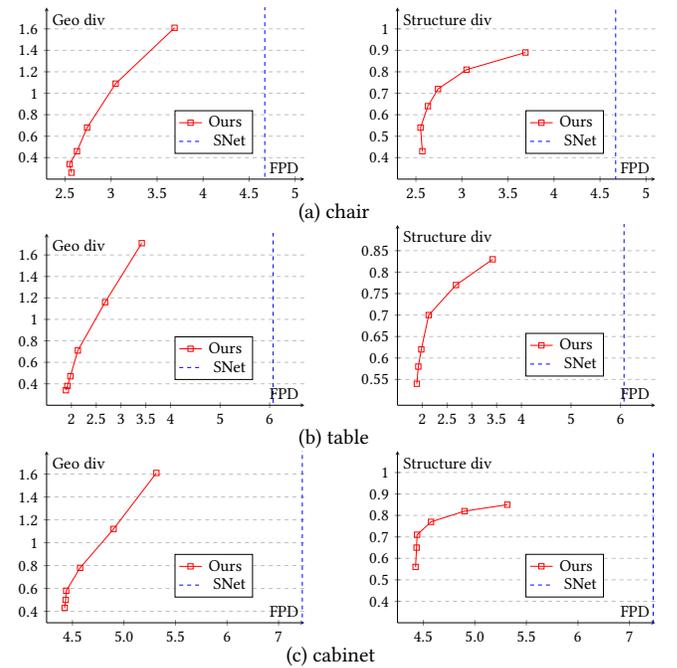

\begin{figure*}
    \centering
  \includegraphics[width=\linewidth]{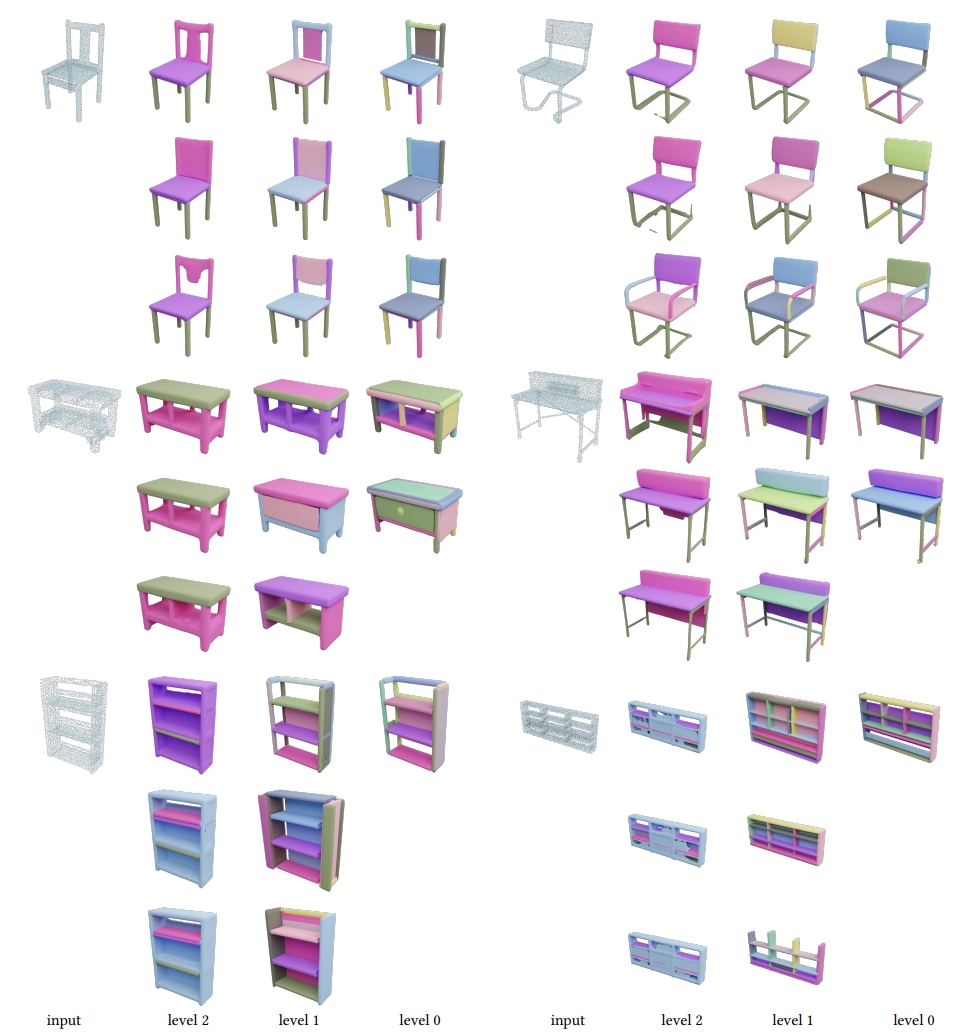}
  \vspace{-12pt}

    \caption{
    \textbf{Structured shape generations}. For each input point cloud, we encode it into a root node and apply the downward rewriting process interleaved with shape code perturbation to generate three different shapes in rows. 
    Moving from left to right, the part nodes are perturbed and progressively refined, depicting the transition from the root level to finer structures at leaf level, following the hierarchy (level 2 \(\rightarrow\) 0). Since different shapes have varying numbers of hierarchical levels, the examples cannot be aligned to the right. The last subfigure in each row shows the final generation result with fine-grained structures (\ie leaf nodes).
    }
    \label{fig:vis_gen_div_chair}
\end{figure*}

\subsubsection{Shape-conditioned generation}\label{sec:gen_shape_cond}

\paragraph{Generation scheme} 
Given the point cloud of a conditioning shape, we first encode it into a shape code through $E_s(E_p(\cdot))$. 
Starting with this code as root node, we iteratively: 
\begin{enumerate}
    \item perturb the nodes by $\vb{s} \leftarrow \vb{s} +  \vb{n}$, where $\vb{n}{\sim} \mathcal{U}(-\sigma, \sigma)$ samples a uniform noise vector and $\sigma$ specifies the strength; 
    \item rewrite the perturbed nodes into lower-level nodes with more specific details; 
\end{enumerate}
until reaching the leaf nodes, i.e., when further rewriting does not change the node numbers. 
During this process, for each node, the random perturbation within the regularized latent space $\vb{\Sigma}$ enables shape variation, and the generative diversity is accumulated by the top-down rewriting.
Combinations of perturbation and downward rewriting also allow for level-of-detail generation, as illustrated by the different paths of Fig.~\ref{fig:vis_gen_lod}.

For quantitative evaluation, we randomly sample 1000 shapes in the training set for each category, and then for each conditioning shape, we randomly generate 10 shapes by the above mentioned process of interleaved shape code perturbation and downward rewriting.
We compare our method with StructureNet \cite{mo2019structurenet} on generation diversity and quality\footnote{In terms of level-of-detail generation, LSD-StructureNet \cite{RobertsLSD2021} is closer to our work, but unfortunately there is no publicly available code for comparison, and its generation quality is comparable to StructureNet.}.

\paragraph{Generation quality}
We evaluate generation quality in terms of geometry-based Frechet inception distance (\textbf{FPD}), coverage (\textbf{COV}) and minimum matching distance (\textbf{MMD}) \cite{li2021spgan} against the test sets.
In particular, we use the DGCNN \cite{dgcnn} pretrained on ModelNet classification to extract point cloud features, on which to fit Gaussian distributions and compute FPD.
For each GT test object, COV measures the chamfer distance to the closet generated shape, thus lower being better; conversely, for each generated shape, MMD measures the chamfer distance to the closet GT test object, and again lower is better.

As shown in Tab.~\ref{tab:eval_shape_gen}, our generated results show significantly better FPD than results of StructureNet.
Our coverage is also better, despite that StructureNet uses fully random sampling of its latent space.
Meanwhile, the MMD of StructureNet is only slightly better than our results, meaning its generated shapes are marginally closer to test set instances on average.
Visual results in Fig.~\ref{fig:vis_gen_div_chair} show that our generated shapes have clean structures and surfaces, in contrast to StructureNet results with detached structures and discrete points.

Additionally, we visualize the generation results at varying noise levels to provide a more comprehensive illustration of the impact of different noise levels. 
Please refer to Fig. 7, Fig. 12 and Table 6 in the supplemental for more evaluation and visualization results.

\paragraph{Generation diversity}
Following \cite{RobertsLSD2021}, we measure the diversity of each group of 10 shapes by computing the mutual chamfer distance for geometric diversity, and counting the different hierarchical structures for structure diversity.
As shown in Fig.~\ref{fig:eval_gen_div}, with increased strength of shape code perturbation, the diversity increases as well; however, all the variations have significantly better shape faithfulness than StructureNet results, as reflected in FPD.

\subsection{Result analysis}
\label{subsec:result_analysis}

In this part we analyze the generalization ability of the trained network through its performance on small categories of PartNet and on multiple objects.
We also analyze the effect of probabilistic sampling for modeling multiple structures.

\begin{figure}
    \centering
  \includegraphics[width=\linewidth]{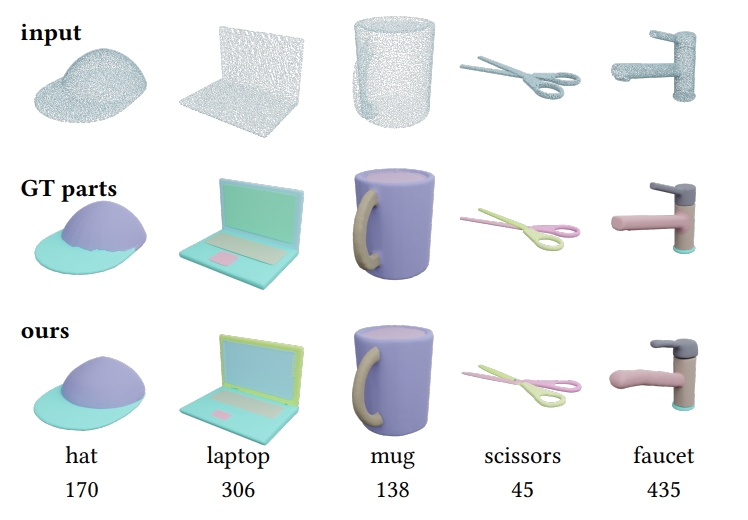}
  \vspace{-24pt}
    
    \caption{\textbf{Qualitative results on small categories}. Our model generalizes across categories and works well even for classes with very few training samples (numbers of training samples are given beneath the class names).}
    \label{fig:small_categories}
\end{figure}
\begin{table}[t]
\fontsize{8}{12}\selectfont
  \centering
  \caption{ \textbf{Quantitative evaluation of small categories on shape reconstruction}. The numbers of training and testing samples are given. The overall performance is on par with big categories (Tab.~\ref{tab:reconeval}).
  }
    \begin{tabular}{ccccccc}
    \toprule[1pt]        
    \textbf{Category} & \textbf{\#Train} & \textbf{\#Test} & \textbf{CD} $\downarrow$ & \textbf{F1} $\uparrow$ & \textbf{Hier} $\downarrow$\\
    \midrule

    Hat     & 170 & 45 & 0.51  & 0.93  & 0.17 \\
    Laptop  & 306 & 82 & 0.25  & 0.98  & 0.15 \\
    Mug     & 138 & 35 & 0.58  & 0.90  & 0.39 \\
    Scissors & 45 & 13 & 0.17  & 0.97  & 0.18 \\
    Faucet  & 435& 134 & 0.51  & 0.90  & 0.51 \\

    \bottomrule[1pt]
    \end{tabular}%
  \label{tab:reconeval_small_cat}%
  \vspace{0mm}
\end{table}%

\begin{figure}
    \centering
  \includegraphics[width=\linewidth]{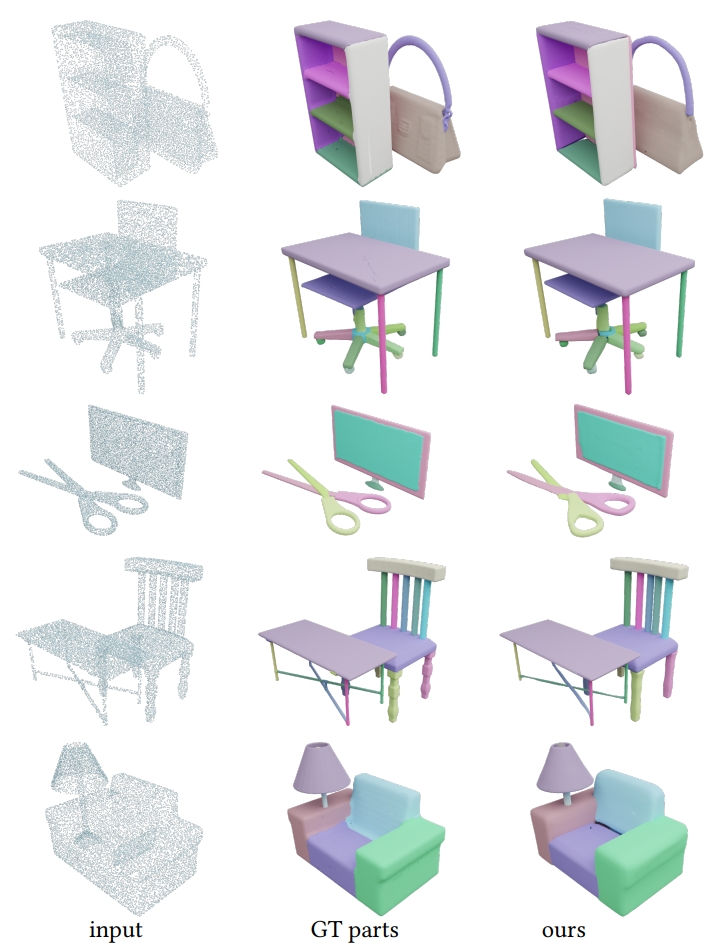}
  \vspace{-24pt}
    
    \caption{\textbf{Rewriting multiple objects}. The input point clouds casually combine pairs of objects. Due to the local rewriting approach, our network generalizes to multiple objects by processing them simultaneously. }
    \label{fig:multiple_objects}
\end{figure}

\begin{figure*}[h]
    \centering
  \includegraphics[width=0.98\linewidth]{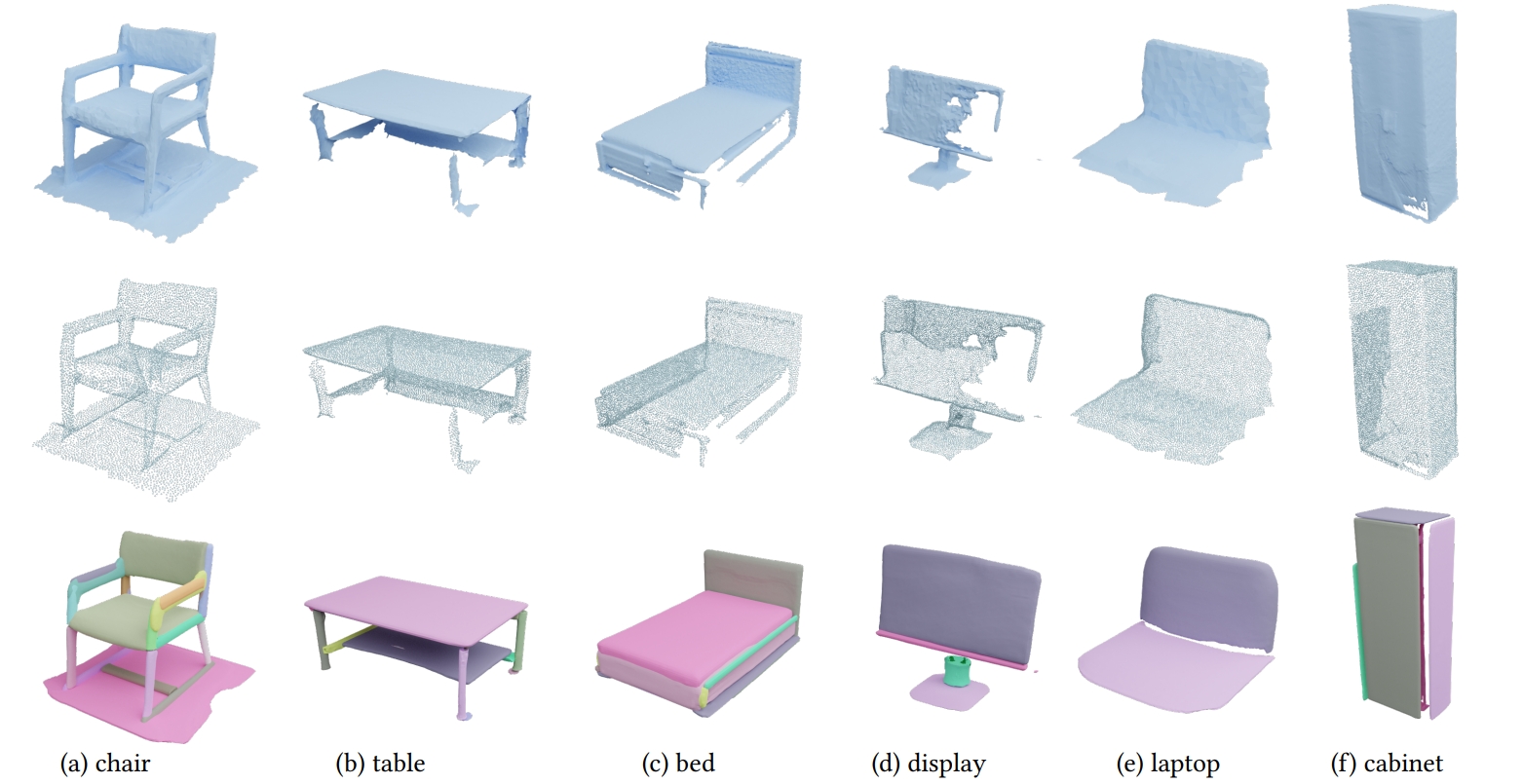}
  \vspace{-12pt}
    \caption{Tests on real data (ScanNet \cite{dai2017scannet}). 
    The three rows show: (1) the object mesh cropped from the scene mesh in ScanNet, (2) the sampled point cloud used as model input, and (3) the model predictions.
    Our model demonstrates robust performance on real-world data across various categories and levels of missing information (a\(\rightarrow\)d).  However, when the missing ratio is large, as shown in (e) and (f), mis-predictions can occur. 
    In (e), the laptop is mis-predicted as chair-like parts, and in (f), the missing section of the cabinet is not accurately reconstructed. 
    This is due to our model not being trained on partial data, making it challenging to differentiate between complete and incomplete shapes when a significant portion of the data is missing. 
    }
    \label{fig:real_scan_scannet}
\end{figure*}

\paragraph{Real data}
We test our method on real-world scans from the ScanNet dataset \cite{dai2017scannet}, as shown in Fig. \ref{fig:real_scan_scannet}. We present six test cases from different categories, each cropped from a scanned indoor scene. It is worth noting that real-world data often includes noise and incomplete regions. Although our model was not trained on noisy or partial data, it still demonstrates robustness in handling such scenarios. However, as the missing ratio increases, structural ambiguity becomes more pronounced, and our method struggles with these cases, as illustrated in Fig. \ref{fig:real_scan_scannet} (e) and (f).

\begin{figure*}
    \centering
  \includegraphics[width=\linewidth]{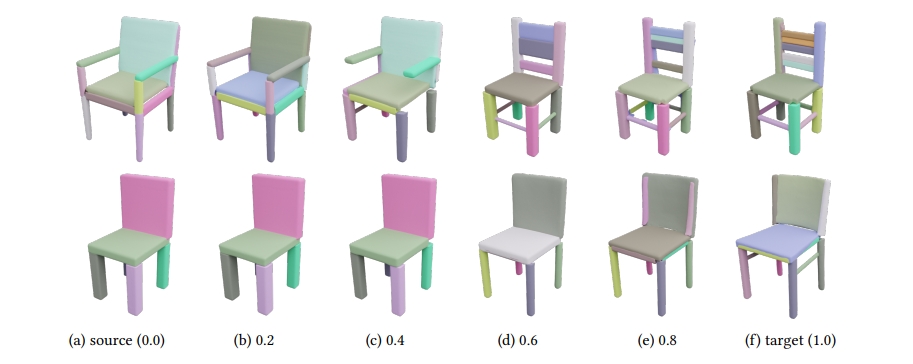}
  \vspace{-20pt}

    \caption{
    \textbf{Shape interpolation} between the source shapes (a) and the target shapes (f). Source and target shapes are first encoded into shape codes. A linear blending weight is used to interpolate between a pair of source and target shape codes. The interpolated shape codes are expanded into detailed structures by our rewriting model in a top-down manner. The intermediate shapes (\(b \rightarrow e\)) represent smooth transitions between the source and target shapes, with the numbers indicating the blending weight.}
    \label{fig:interp}
\end{figure*}

\paragraph{Shape interpolation}
To assess the quality of the learned latent space, we inspect the smoothness of transitions for shape interpolation. Given a source shape \(S\) and a target shape \(T\), we first encode the two shapes using our patch and part encoder to get their shape codes $\vb{z}_S,\vb{z}_T$. Then, a linear blending weight \(\alpha \in [0,1]\) is used to interpolate between the two shape codes \(\vb{z}_{\alpha}=\alpha \vb{z}_T + (1 - \alpha) \vb{z}_S\). The interpolated codes are then decoded into structured shapes via our generation mechanism. Fig. \ref{fig:interp} presents two interpolation examples, where the intermediate shapes (b)$\rightarrow$(e) transit smoothly in both structure and geometry between the source and target shapes.

\paragraph{Generalization}

For small categories without sufficient data samples, training specific models can easily overfit, while our model generalizes to such categories with robustness.
Indeed, as shown in Fig.~\ref{fig:small_categories} and Tab.~\ref{tab:reconeval_small_cat}, the reconstruction accuracy on these small categories are on par with those large categories with abundant training data.
Similar results for remaining categories are reported in the supplemental.

Multiple object combinations can be robustly handled as well, as shown in Fig.~\ref{fig:multiple_objects}.
Such generalization is facilitated by both our rewriting mechanism that handles local part compositions and the data augmentation that randomly combines shape parts to prepare the network for flexibility.
Note that the input point clouds in Fig.~\ref{fig:multiple_objects} have combinations that are never seen in training data.

\begin{figure*}[tb]
    \centering
  \includegraphics[width=0.98\linewidth]{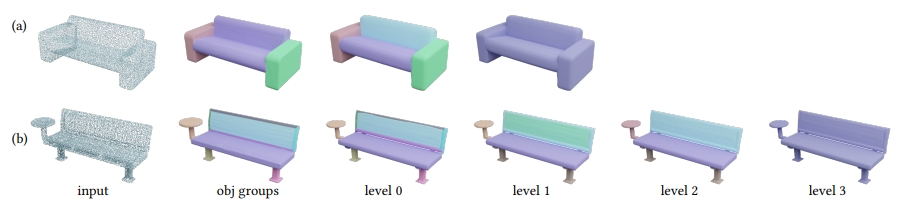}
  \vspace{-10pt}
	\caption{\textbf{Hierarchy on ShapeNet objects not contained in PartNet}. Although ShapeNet objects are not trained with hierarchy supervision, the trained network generalizes to those resembling PartNet categories in zero-shot and produces plausible hierarchies.}
	\label{fig:shapenet_hierarchy}
\end{figure*}
\begin{figure}
	\centering
  \includegraphics[width=\linewidth]{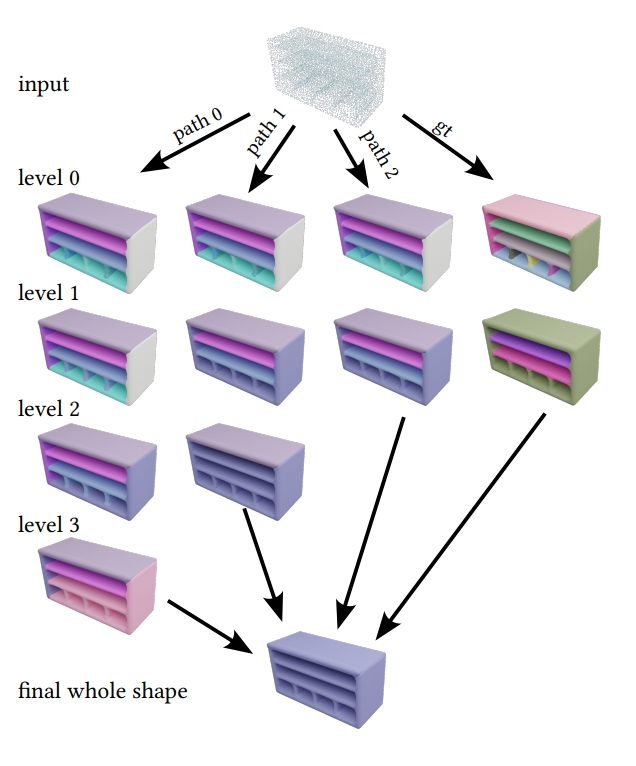}
  \vspace{-20pt}
     \caption{\textbf{Sampling different hierarchical structures}. Starting from input, by iteratively applying the rewriting model with random sampling, we obtain three paths encoding hierarchies that are all different from the GT annotation, showing diversity that is vital for addressing ambiguity. }
	\label{fig:sampling}
\end{figure}

\paragraph{Probabilistic sampling}
Within each rewriting step, the probabilistic sampling of decoder outputs is used for resolving structure ambiguity during training and supports the generation of different structures during testing (Sec.~\ref{subsec:sampling}).
In particular, during the testing stage, we follow a schedule $[\delta_t], t\in [T]$ to conduct confidence guided sampling.
By using a strict schedule where $\delta$ starts from being close to 1 and decreases gradually, we obtain samples which have close to optimal probabilities; on the other hand, by using a relaxed schedule where $\delta$ stays at a permissive level (e.g. around 0.5), the samples become more diverse.
For evaluations in other sections, we have used a strict linear schedule of $\delta_t = 1 - 0.1t$ to sample the more likely solutions.
In this section, we use a relaxed schedule of $\delta_t = 0.3$ to show more diverse sampling results.

As shown in Fig.~\ref{fig:sampling}, given an input shape depicting a cabinet, there can be multiple paths transitioning from the unstructured input to structured parts of different levels.
Such paths are obtained by iteratively applying the rewriting model, and in each iteration using a random sampling of decoded parts.

\subsection{ShapeNet extension}
\label{subsec:shapenet}

\begin{figure}[tb]
    \centering
  \includegraphics[width=\linewidth]{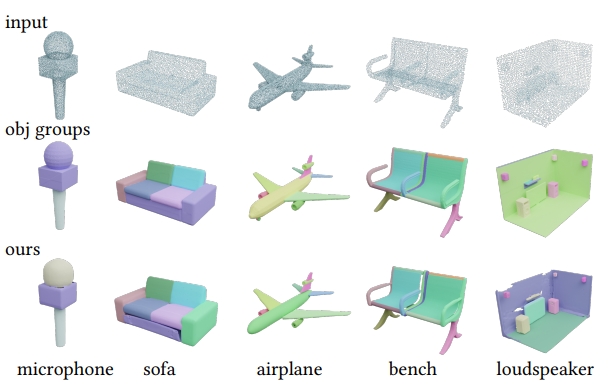}
  \vspace{-20pt}
    \caption{\textbf{Results on ShapeNet objects not contained in PartNet}. The ShapeNet GT parts are obtained by heuristics of obj file groups and connected components and not always plausible (e.g., microphone and loudspeaker). Nonetheless our model recovers meaningful parts by leveraging rules learned from large scale dataset across categories.}
    \label{fig:shapenet_parts} \vspace{0mm}
\end{figure}

PartNet represents a large-scale and systematic manual effort on annotating the hierarchical structures of a portion of ShapeNet models.
For the rest of ShapeNet data, there are generally no available annotations to train and evaluate model performances except for few classes \cite{sun2022semi}.
Nevertheless, we have tried to extend our model to ShapeNet data by training on leaf nodes that are roughly constructed by obj groups and connected components.
We inspect how the model works on these testing data, and how few-shot tuning can quickly adapt the model.

First, we note that our model can produce more plausible and consistent leaf nodes from raw input for ShapeNet objects (Fig.~\ref{fig:shapenet_parts}), which can be attributed to the strong regularization effect of a single trained model on heuristic annotations of different qualities.
A comprehensive report of leaf node reconstruction for ShapeNet is given in the supplementary.
In addition, our model constructs hierarchies for ShapeNet objects that resemble PartNet categories, which is a zero-shot transfer of hierarchies learned from PartNet (Fig.~\ref{fig:shapenet_hierarchy}).

\begin{table}[t]
\fontsize{8}{12}\selectfont
  \centering
  \caption{\textbf{Few-shot tuning}. Samples denotes the number of samples for fine-tuning. Iter gives the number of fine-tune iterations that lead to convergence on validation set. CD is squared chamfer distance of scale $\times 10^{-2}$. F-score has been omitted due to the results all having near perfect values. Hier denotes the tree editing distance of reconstructed hierarchies from annotated GT. }
    \begin{tabular}{cccccccc}
    \toprule[1pt]        
    \textbf{Samples} & \textbf{Iter} & \textbf{CD} $\downarrow$ & \textbf{Hier} $\downarrow$\\
    \midrule
    0           & 0     & 0.11  & 1.01 \\
    10	        & 3k	& 0.14	& 0.30 \\
    50	        & 6k	& 0.12	& 0.12 \\
    100	        & 12k	& 0.12	& 0.12 \\
    1281(ALL)   & 160k	& 0.11	& 0.08 \\

    \bottomrule[1pt]
    \end{tabular}%
  \label{tab:few_shot}%
  \vspace{0mm}
\end{table}%
\begin{figure}
    \centering
  \includegraphics[width=\linewidth]{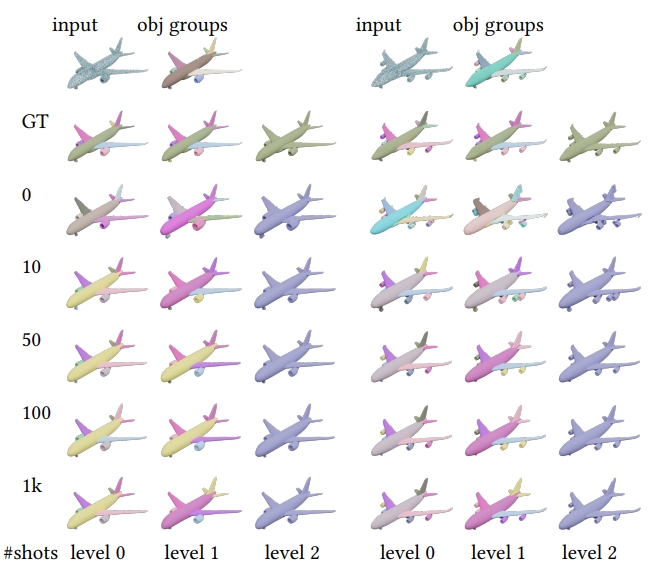}
  \vspace{-20pt}
    
    \caption{\textbf{Few-shot finetuning} on airplanes annotated by \cite{sun2022semi}. The zero-shot test results frequently have scattered parts due to the training on irregular obj group partitions. As more samples are used for finetuning $(E_R,D_R)$, the predictions quickly adapt to the new GTs. }
    \label{fig:vis_recon_few_shot}
\end{figure}

For objects whose hierarchies are significantly out-of-distribution from PartNet annotations, we resort to few-shot tuning of the model.
A pretrained model capable of generalization can be quickly adapted to new data distributions by finetuing with few data samples \cite{FewShotSurvey23}. 
To validate if our model has this property, we finetune it on samples with hierarchical annotations of airplanes provided by \cite{sun2022semi}.
In particular, we only update each of the last transformer layers of $E_R$ and $D_R$, and freeze the other parameters. In each batch, we combine in equal proportion the novel samples (and their augmentations, Sec.~\ref{subsec:augmentation}) with random samples from previous training data, and use a small $3\times 10^{-5}$ learning rate.
The finetuned model is tested on the testing split from \cite{sun2022semi}.
We evaluate the reconstruction performance of the finetuned model (Sec.~\ref{subsec:comparison_recon}).

As shown in Tab.~\ref{tab:few_shot}, it is expected that the zero-shot results are not compatible with the new data and have low hierarchy accuracy, due to the new data having much coarser parts than the material group labels we have trained on (Fig.~\ref{fig:vis_recon_few_shot}).
But beyond that, 50 samples and 6k iterations can already adapt the pretrained network to quality predictions on the new test set, comparable to using the full 1k training samples.
From Fig.~\ref{fig:vis_recon_few_shot}, we can see that as the number of samples goes up, the finetuned network transitions from the scattered predictions induced by obj grouping to the labeled structures of \cite{sun2022semi} smoothly.
Such results demonstrate that our pretrained model has good generalization potential and can be easily adapted by few-shot finetuing.

\subsection{More modeling applications}
\label{subsec:applications}

\begin{figure}
    \centering
  \includegraphics[width=\linewidth]{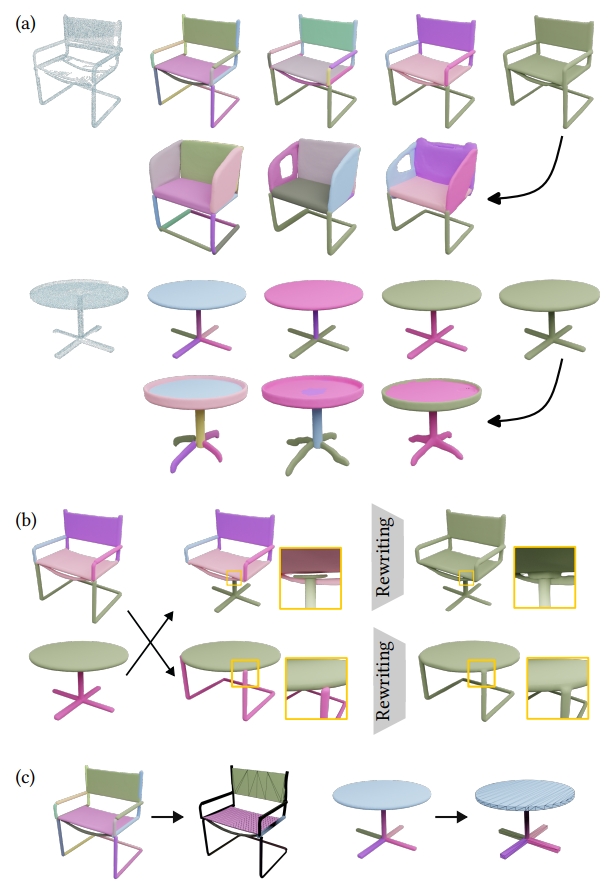}
  \vspace{-12pt}
    \vspace{-3mm}
    \caption{ \textbf{Modeling applications enabled by rewriting}. 
    \textbf{(a)} reconstruction plus variation. Given partial scans, our rewriting model parses them into structured hierarchies, and explores variations by downward rewriting plus shape perturbation. 
    \textbf{(b)} part editing. By selecting parts at proper levels for objects of different categories, swapping them, and rewriting upward to regularize the editing, novel objects are obtained.
    \textbf{(c)} shape vectorization. After a shape is parsed into parts, nearest parts with compact polygonal meshes can be retrieved from a database to vectorize the shape.
    }
    
    \label{fig:applications}
    \vspace{0mm}
\end{figure}

In this section, we demonstrate some of the application scenarios that are enabled by the structure rewriting scheme, including reconstruction followed by structure aware generation, rewriting for regularized editing, and vectorization by part mesh retrieval.

\paragraph{Structured reconstruction plus generation}
Given a partial scan, we can apply upward rewriting to convert and complete it into structured shapes, and further explore the variations of this shape by downward rewriting plus shape code perturbation (Sec.~\ref{subsec:comparison_gen}). 
Examples are shown in Figs.~\ref{fig:teaser} and \ref{fig:applications}(a).

\paragraph{Part-level editing}
Given objects represented by explicit parts, we can directly edit them by manipulating parts, and further regularize the editing through rewriting.
For example, as shown in Fig.~\ref{fig:applications}(b), first we swap the bases of a chair and a table by interchanging their corresponding parts, which produces gaps at junctions.
To fix these gaps, we simply rewrite them upward where those edited parts are merged with other parts seamlessly.
Currently, we manually swap parts for each editing operation. Extending our method to learn an editing latent space, similar to StructEdit \cite{Mo2020StructEdit}, could help automate and speed up the editing process, especially for larger-scale operations.
Please refer to the supplemental for more editing details and examples.

\paragraph{Shape vectorization by part-based mesh retrieval}
Given the structured reconstruction into parts, our encoding of parts via latent codes permits querying similar parts from the training set, whose meshes can then replace the marching cube results to vectorize the corresponding parts with compactness.
The nearest neighbor query of similar parts is done by simple cosine similarity computation: given a reconstructed part with shape code $\vb{s}$, and a set of training shape codes $[\vb{s}_i]$, we find the closet part by $\argmax_i \frac{\vb{s}\cdot \vb{s}_i}{\|\vb{s}\| \|\vb{s}_i\|}$.
Two examples are shown in Fig.~\ref{fig:applications}(c).

\subsection{Limitations and future work}
\label{subsec:limitation}

\begin{figure}
    \centering
  \includegraphics[width=\linewidth]{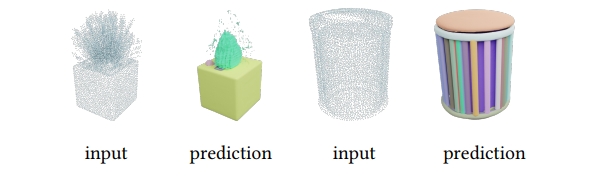}
  \vspace{-20pt}
    \caption{
    \textbf{Failure cases}. 
    On the left, the many details of the plant in vase are not well captured by a single part. On the right, the trashcan has regularly distributed bars that are not evenly positioned in the reconstructed model.
    }
    \label{fig:vis_failure_cases}
\end{figure}

Our work only makes a first step toward rewriting based structured shape modeling. 
However, for shapes with complex geometry, such as the vase in Fig. \ref{fig:vis_failure_cases} left, our model struggles to reconstruct accurate details. Additionally, for shapes with regular patterns, such as the trashcan in Fig. \ref{fig:vis_failure_cases} right, the model has difficulty recovering the patterns precisely. 
To enhance the network accuracy in these cases, one can use finer patch grids (e.g. $16^3$) to recognize small parts, learn part rotations for more compact shape codes, and detect and apply regular patterns as constraints.
Additionally, there are several major directions worth exploration in future works. 

\paragraph{Enriching semantics}
Compared with previous category specific works, our network does not explicitly encode the semantics of object parts or the symmetry among sibling parts, but rather focus on cross-category and novel category generalization. 
However, given larger-scale data, both semantics and symmetrical constraints can be encoded by augmenting parts with additional attributes ($\Sigma$ of Eq.~(\ref{eq:rewrite})) and the rewriting network with corresponding embedding and prediction layers (Sec.~\ref{subsec:rewrite_network}).

\paragraph{Self-supervised and weakly supervised learning}
Toward greater generalization, it is desirable to extend the current framework for self-supervised learning or weak supervision from pretrained large image models.
For self-supervised learning, the chaining of three stages ($D_p\circ D_s\circ D_R\circ E_R\circ E_s\circ E_p$) and two directions ($\uparrow,\downarrow$) form closed loops for reconstruction based self-supervision on novel training shapes.
For weakly supervised learning, pretrained image foundation models \cite{SAM2023} provide hierarchical part labels that can be aggregated for 3D supervision \cite{liu2023partslip}.

\paragraph{General rewriting tasks}
Rewriting models can capture semantics broader than the top/down traversal of shape part hierarchies \cite{Dershowitz1990RewriteS}. 
By extending the direction commands with task specifications like examples, user interactions and textual instructions \cite{InstructGPT,ICLTaskVector2023}, it is possible to extend the rewriting model to more general conditioned generative modeling that targets function and affordance reasoning based on structured shapes, which is critical for robotics and embodied artificial intelligence \cite{VisualAffordanceSurvey}.

\section{Conclusion}
We have presented a novel \name framework for structured shape modeling through rewriting.
Contrary to previous works that address the ambiguity of structured representations by relying on categorical templates, the rewriting model learns local and probabilistic rewriting rules that accommodate ambiguous decompositions and enable cross-category generalization.
Building on regular latent spaces for geometry patches and part shapes, the rewriting model with simple syntax can be learned with a pair of transformer encoder and decoder that supports sampling through iterative decoding.
Being local and probabilistic, the model can be trained by synthesized large-scale combinatorial data to enhance its generalization.
Experiments show that \name achieves superior performances on structured shape reconstruction and generation, and generalizes to small and unseen categories by zero shot or few-shot tuning, as well as scenes of multiple objects.
The chaining of multiple rewriting steps also permits novel modeling scenarios.

\bibliographystyle{ACM-Reference-Format}
\bibliography{src/rewrite}

\end{document}